\documentclass{bmvc2k}

\usepackage{pdfpages} 

\title{One-Pot Multi-Frame Denoising}

\addauthor{Lujia Jin}{jinlujia@pku.edu.cn}{123}
\addauthor{Shi Zhao}{magishe@pku.edu.cn}{4}
\addauthor{Lei Zhu}{zhulei@stu.pku.edu.cn}{123}
\addauthor{Qian Chen}{chen_qian@stu.pku.edu.cn}{123}
\addauthor{Yanye Lu\textsuperscript{\Envelope}}{yanye.lu@pku.edu.cn}{235}

\addinstitution{
 Department of Biomedical Engineering,\\
 Peking University, Beijing, China
}
\addinstitution{
	Institute of Medical Technology,\\
	Peking University, Beijing, China
}
\addinstitution{
	Institute of Biomedical Engineering,\\
	Peking University Shenzhen Graduate School, Shenzhen, China
}
\addinstitution{
	School of Physics,\\
	Peking University, Beijing, China
}
\addinstitution{
	National Biomedical Imaging Center,\\
	Peking University, Beijing, China
}

\runninghead{Jin et al.}{One-Pot Multi-Frame Denoising}

\begin{document}

\maketitle
\vspace{-1.5ex}
\begin{abstract}
The performance of learning-based denoising largely depends on clean supervision. However, it is difficult to obtain clean images in many scenes. On the contrary, the capture of multiple noisy frames for the same field of view is available and often natural in real life. Therefore, it is necessary to avoid the restriction of clean labels and make full use of noisy data for model training. So we propose an unsupervised learning strategy named one-pot denoising (OPD) for multi-frame images. OPD is the first proposed unsupervised multi-frame denoising (MFD) method. Different from the traditional supervision schemes including both supervised Noise2Clean (N2C) and unsupervised Noise2Noise (N2N), OPD executes mutual supervision among all of the multiple frames, which gives learning more diversity of supervision and allows models to mine deeper into the correlation among frames. N2N has also been proved to be actually a simplified case of the proposed OPD. From the perspectives of data allocation and loss function, two specific implementations, random coupling (RC) and alienation loss (AL), are respectively provided to accomplish OPD during model training. In practice, our experiments demonstrate that OPD behaves as the SOTA unsupervised denoising method and is comparable to supervised N2C methods for synthetic Gaussian and Poisson noise, and real-world optical coherence tomography (OCT) speckle noise.
\vspace{-1.5ex}
\end{abstract}

\begin{figure}[t]
	\centering
	\includegraphics[height=4.2cm]{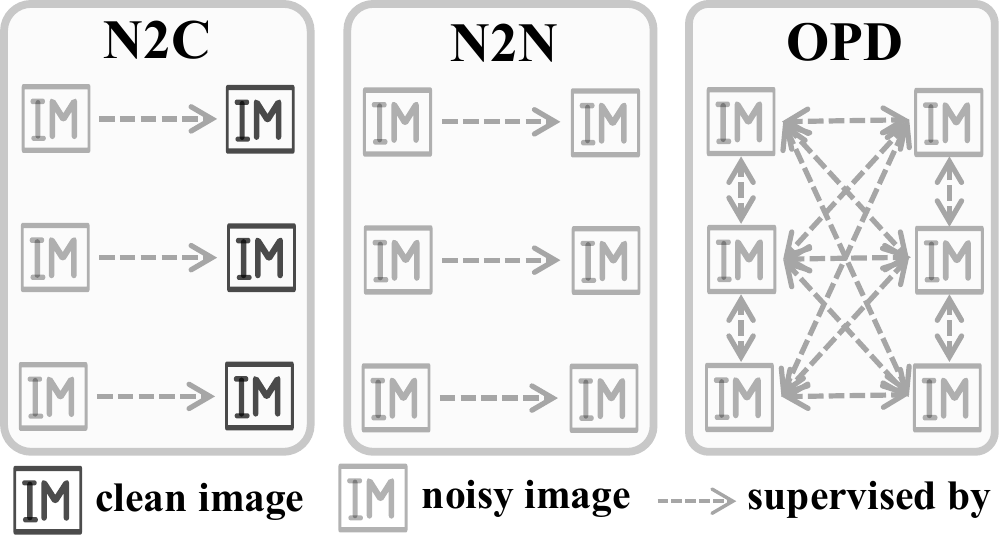}
	\caption{Comparison between our OPD and other supervision strategies, including supervised N2C and unsupervised N2N. The OPD establishes mutual supervision among multiple frames, which allows the hidden clean image to be better estimated.}
	\label{fig:1}
	\vspace{-1.5ex}
\end{figure}

\section{Introduction}
Due to the non-definite nature of image denoising, it is always difficult for methods based on reasoning to perform as well as expected on a lot of scenes. The turning milestone appears in the rise of deep learning, which greatly develops image denoising and meanwhile ignites the need for data. Convolutional neural networks (CNN) with various structures and characteristics have been designed \cite{zhang2017beyond,guo2019toward,lefkimmiatis2018universal,tian2020attention,anwar2019real}. They do not care about the causal inference of noise pattern but learn end-to-end from the noise image to its clean counterpart. However, clean images are difficult to obtain, making conventional learning under Noise2Clean (N2C) almost impossible to proceed. Noise2Noise (N2N)\cite{lehtinen2018noise2noise} overcomes this obstacle at the cost of one more noisy image. The noisy-clean image pair is replaced by the two paired noisy images to train the model under the N2N strategy.

Although it is hard to get a clean image in real-life scenes, in many cases the acquisition of multi-frame noisy images is available and even natural, such as exposure bracketing\cite{bertalmio2009fusion}, astrophotography\cite{guilloteau2020hyperspectral}, and optical coherence tomography (OCT)\cite{qiu2020noise}, etc. Denoising for these scenes is called multi-frame denoising (MFD), which aims at finding a mapping $\bm{f}$ with a given multi-frame dataset $\left\{\left(\bm{x}_{i}+\bm{n}_{i}^{j},\bm{y}_{i}\right)|i\in\left[1,N\right],j\in\left[1,m\right]\right\}$ such that $\bm{f}\left[\bigcup_{j\in\left[1,m\right]}\left(\bm{x}_{i}+\right.\right.\\
\left.\left.\bm{n}_{i}^{j}\right)\right]=\bm{y}_{i},i\in\left[1,N\right]$. As can be seen from the above problem statement, MFD is essentially a reasonable fusion of multiple noisy frames corresponding to the same underlying clean reference. All previous MFD methods\cite{mildenhall2018burst,marinvc2019multi,xia2020basis} are supervised methods. For the first time, we propose an unsupervised strategy, one-pot denoising (OPD), to achieve high-quality denoising of multiple frames. As demonstrated in Fig. \ref{fig:1}, OPD strives to extend the supervision from noisy-clean pairs to a group of all noisy frames corresponding to the same clean target. Specifically, we design a data allocation method that continuously shuffles the supervision pair from multiple frames during training so that the information contained in each frame could be fully utilized. This implementation is named OPD-random coupling (OPD-RC). In addition, we construct an alienation loss function, which has an equivalent denoising effect to OPD-RC and provide a more formulaic and therefore more intuitive understanding for the OPD strategy. This implementation is named OPD-alienation loss (OPD-AL). Both implementations of OPD are experimented on additive white Gaussian noise (AWGN), signal-dependent Poisson noise, and OCT speckle noise. Experimental results on all three noises show that, both qualitatively and quantitatively, OPD performs better than N2N and sometimes outperforms N2C.

In a nutshell, the main contributions of our work are as follows:

\begin{enumerate}
	\vspace{-1.5ex}
	\item  We propose OPD, a denoising strategy based on an unprecedented mutual supervision paradigm. OPD is the first proposed unsupervised MFD method.
	\vspace{-1.5ex}
	\item  From the perspectives of data allocation and loss function, two specific implementations, OPD-RC and OPD-AL are presented for MFD. We also reveal that the well-known N2N\cite{lehtinen2018noise2noise} can be interpreted as a simplified case of our OPD.
	\vspace{-1.5ex}
	\item Experiments show that our OPD behaves as the SOTA unsupervised denoising method and is comparable to supervised N2C methods for several MFD tasks, including denoising AWGN and Poisson noise, and OCT speckle noise reduction.
\end{enumerate}  

\vspace{-1.ex}
\section{Related Works}
\vspace{-1.ex}
\subsection{Multi-frame Denoising}
Compared to single-image denoising (SID), MFD has received significantly less attention in the past decade. Tico\cite{tico2008multi} first migrated the popular Non-Local Means (NLM)\cite{buades2005non} from single image denoising to MFD. This method compares the similarity of blocks not only within but also among frames. V-BM3D\cite{kostadin2007video} and V-BM4D\cite{maggioni2011video,maggioni2012video} are based on the famous BM3D\cite{dabov2007image} to denoise videos through sparse 3D transform-domain collaborative filtering. Buades et al. \cite{buades2009note} provided a complex processing chain including accurate registration and noise estimation. Hasinoff et al.\cite{hasinoff2016burst} applied an FFT-based alignment algorithm and a hybrid 2D/3D Wiener filter to burst denoising, which had been built atop Android's Camera2 API. Completely different from the aforementioned block-based fusion methods, accumulation after registration (AAR)\cite{buades2010multi} directly uses weighted averaging to fuse multiple frames, which is proved to be effective in zero-mean noise reduction. 

All of the above are non-learning methods, while learning-based methods are rarer. Godard et al.\cite{godard2018deep} constructed a simple but effective recurrent neural network inspired by series data processing. Mildenhall et al.\cite{mildenhall2018burst} proposed a kernel prediction network (KPN) to produce clean images from bursts by unique 3D denoising kernels. Marinc et al.\cite{marinvc2019multi} used multi-scale kernels to extend KPN to multi-KPN (MKPN). Furthermore, Xia et al.\cite{xia2020basis} developed a basis prediction network (BPN) for effective burst denoising with large kernels, which achieves both significant quality improvement and faster run-time.
\vspace{-1.ex}
\subsection{Supervision Strategies for Image Denoising}
Most learning-based image denoising methods follow the conventional N2C supervision paradigm\cite{zhang2017beyond,guo2019toward,lefkimmiatis2018universal,tian2020attention,anwar2019real}, which makes it indispensable to obtain clean images as labels. N2N\cite{lehtinen2018noise2noise} breaks this limitation by exploring an alternative supervision paradigm, in which pairs of noisy images corresponding to a common unknown clean target are used for training. Wu et al.\cite{wu2019consensus} showed that the results of optimization are equivalent under N2N and N2C as long as the amount of data is large enough. AltN2N\cite{calvarons2021improved} improves the performance of N2N under limited data by fine tuning. In the past two years, N2N has been widely used in low dose computed tomography\cite{hasan2020hybrid}, positron emmision tomography\cite{kang2021noise2noise}, synthetic aperture rader\cite{dalsasso2021sar2sar} and other image denoising tasks. Speech denoising\cite{kashyap2021speech}, video enhancement\cite{boiko2020single,zach2020real}, nanochennel measurement\cite{takaai2021unsupervised} and other non-image denoising has also excellently proceeded under N2N.

Going further than N2N, self-supervised strategies denoise with only a single noisy image, and they can be divided into two categories. The first category of methods utilizes priori noise models as an external aid to construct another noisy image and pairs it with the provided data for further N2N learning. Noiser2Noise\cite{moran2020noisier2noise} and Noise-as-Clean\cite{xu2020noisy} are representative methods in this category. These methods are strictly limited by prior noise knowledge. The second category of methods does not require any additional knowledge. Methods based on mask prediction, such as DIP\cite{ulyanov2018deep}, Noise2Void (N2V)\cite{krull2019noise2void}, Noise2Self (N2S)\cite{batson2019noise2self} and blind-splot network\cite{laine2019high}, are typical representatives of this category, but their denoising performance is inferior to the aforementioned prior knowledge-assisted methods.
\vspace{-1.ex}
\section{Methods}\label{sec:3}
In this Section, we first retrospect and formulate three conventional strategies for MFD as preliminaries for subsequent method development (Sec. \ref{sec:3.1}). Then the principle of our OPD and two specific implementations, OPD-RC and OPD-AL, are introduced (Sec. \ref{sec:3.2}). Finally, we compare OPD with other strategies to reasonably analyze their pros and cons (Sec. \ref{sec:3.3}).
\vspace{-1.ex}
\subsection{Retrospecting Conventional MFD}\label{sec:3.1}
MFD aims to learn the underlying clean target $\bm{x}_{i}$ based on the multiple noisy images $\mathbb{X}:\left\{\bm{x}_{i}+\bm{n}_{i}^{j}\;|\;i\in\left[1,N\right],\;j\in\left[1,m\right]\right\}$, where $N$ denotes the number of samples and $m$ refers to the number of noisy frames per sample.

From the perspective of supervision strategies, there are mainly three existing strategies for MFD: AAR\cite{buades2010multi}, N2C and N2N\cite{lehtinen2018noise2noise}, which are described in detail as follows.

\noindent\textbf{AAR for MFD: }As one of the most classical methods, AAR\cite{buades2010multi} is a simple but effective algorithm. After registering all frames corresponding to the same sample, a clean $\bm{x}_{i}$ can be estimated simply by accumulating and averaging as:

\begin{shrinkeq}{-1.5ex}
	\begin{align}
		\widehat{\bm{x}_{i}}=\frac{1}{m}\sum_{j=1}^{m}(\bm{x}_{i}+\bm{n}_{i}^{j}),\quad i=1,2,...,N
		\label{eq:1}
	\end{align}
\end{shrinkeq}

\noindent\textbf{N2C for MFD: }Learning-based methods improve the generalization potential. Based on the clean estimation of AAR, a set of training pairs $\mathbb{T}_{N2C}:\left\{(\bm{x}_{i}+\bm{n}_{i}^{j},\; \widehat{\bm{x}_{i}})\;|\right.$ $\left.\;i\in\left[1,N\right],\;j\in\left[1,m\right]\right\}$ can be built and N2C learning can be performed with the loss:

\begin{shrinkeq}{-1.5ex}
	\begin{align}
		\mathcal{L}_{N2C}=\frac{1}{N\times m}\sum_{i=1}^{N}\sum_{j=1}^{m}\left\|f_{\bm{\Theta}}(\bm{x}_{i}+\bm{n}_{i}^{j})-\widehat{\bm{x}_{i}}\right\|_{2}^{2},
		\label{eq:2}
	\end{align}
\end{shrinkeq}
where $L_{2}$ error is used by default in our derivation.

\noindent\textbf{N2N for MFD: }Among the strategies that help models escape the constraints of clean supervision, N2N is the most representative. Find a random permutation $\mathbb{I}_{i}$ of the sequence $\mathbb{I}=[1,2,...,m]$ for each $i\in\left[1,N\right]$. $\mathbb{J}_{i}$ and $\mathbb{K}_{i}$ are two sequences obtained by equally dividing $\mathbb{I}_{i}$, which means that $\mathbb{J}_{i}$ and $\mathbb{K}_{i}$ constitute a random uniform partition of $\mathbb{I}$. Note that when $m$ is odd, randomly discard an element in $\mathbb{I}$. Treat the elements in $\mathbb{J}_{i}$ and $\mathbb{K}_{i},\;i\in\left[1,N\right]$ as indexes of frames and equally divide the multi-frame data into two parts:

\begin{shrinkeq}{-1.5ex}
	\begin{align}
		\begin{aligned}
			&\mathbb{X}_{1}:\Big\{\bm{x}_{i}+\bm{n}_{i}^{j}\ |\ i\in\left[1,N\right],\ j\in\mathbb{J}_{i}\Big\}\\ 
			&\mathbb{X}_{2}:\Big\{\bm{x}_{i}+\bm{n}_{i}^{k}\ |\ i\in\left[1,N\right],\ k\in\mathbb{K}_{i}\Big\}
		\end{aligned}
		\label{eq:3}
	\end{align}
\end{shrinkeq}
where it should be noted that $j$ and $k$ are just the elements in $\mathbb{J}_{i}$ and $\mathbb{K}_{i}$ but not the indexes.

Then the elements in $\mathbb{X}_{1}$ and $\mathbb{X}_{2}$ can be paired one-to-one to construct the training set $\mathbb{T}_{N2N}$ and N2N learning can be performed with the loss:

\begin{shrinkeq}{-1.5ex}
	\begin{align}
		\begin{aligned}
			&\mathcal{L}_{N2N}=\frac{1}{N\times\left\lfloor\frac{m}{2}\right\rfloor}\sum_{i=1}^{N}\sum_{\mbox{\scriptsize$\begin{array}{c}j\in\mathbb{J}_{i}\\k\in\mathbb{K}_{i}\end{array}$}}\Big\|f_{\bm{\Theta}}(\bm{x}_{i}+\bm{n}_{i}^{j})-(\bm{x}_{i}+\bm{n}_{i}^{k})\Big\|_{2}^{2},\\
			&s.t.\quad idx(j)=idx(k),
		\end{aligned}
		\label{eq:4}
	\end{align}
\end{shrinkeq}
where $idx(j)$ and $idx(k)$ represent the corresponding indexes of $j$ and $k$ in $\mathbb{J}_{i}$ and $\mathbb{K}_{i}$.

\vspace{-1.ex}
\subsection{One-Pot Multi-Frame Denoising}\label{sec:3.2}
In Eq. (\ref{eq:4}), two ways can be found to use a pair of noisy images, which are employing one to supervise the other and vice versa. Based on this consideration, we propose a concept of "mutual supervision". As shown by the bidirectional arrow in Fig. \ref{fig:1}, the roles of the two noisy images participating in the training under mutual supervision are not absolutely prescribed, but interchanged, entangled and equivalent. Furthermore, for multi-frame scenarios with $m>2$, mutual supervision can be established among all noisy images corresponding to the same $\bm{x}_{i}$. The learning strategy based on the above concept is evocatively named OPD. OPD enables each noisy image to play an equally important role. Diversified samples and labels enable the model to squeeze out much more hidden inter-frame information contained in the data during learning. At the same time, this also makes the model face more optimization possibilities, which potentially leads to an improvement in denoising performance.

The most intuitive way to perform OPD is to go through all the one-to-one unidirectional supervision pairs. Nevertheless, it is easy to realize that skyrocketing data size makes this way so crude. Considering that the learnable models usually look for the minimum in an iterative manner on the hypersurface defined by the loss function, from the perspectives of reconstruction of the iterative data pairs and reconstruction of the loss function, we propose two feasible OPD implementation methods, OPD-RC and OPD-AL, respectively.

\noindent\textbf{OPD-RC: }Since the data is fed into the model iteratively during training, we can simply shuffle the multiple frames each time before a new iteration to continuously reconstruct the supervision direction. Assuming that $s$ refers to a step during model updating, before the $s$th iteration, construct random and uniform partition $\mathbb{J}_{i}^{s}$ and $\mathbb{K}_{i}^{s}$ of the sequence $\mathbb{I}=\left[1,2,...,m\right]$. Then randomly divide the $m$ noisy frames corresponding to a same $\bm{x}_{i}$ into two sets:

\begin{shrinkeq}{-1.5ex}
	\begin{align}
		\begin{aligned}
			&\mathbb{X}_{1}^{s}:\Big\{\bm{x}_{i}+\bm{n}_{i}^{j}|i\in\left[1,N\right],j\in\mathbb{J}_{i}^{s},s\in\mathbb{N}^{*}\Big\}\\ 
			&\mathbb{X}_{2}^{s}:\Big\{\bm{x}_{i}+\bm{n}_{i}^{k}|i\in\left[1,N\right],k\in\mathbb{K}_{i}^{s},s\in\mathbb{N}^{*}\Big\}
		\end{aligned}
		\label{eq:5}
	\end{align}
\end{shrinkeq}

According to the back-propagation rule, the model at step $s+1$ can be updated as:

\begin{shrinkeq}{-1.5ex}
	\begin{align}
		\begin{aligned}
			&\bm{\Theta}^{s+1}=\bm{\Theta}^{s}-\eta\frac{\partial}{\partial\bm{\Theta}}\bigg [\sum_{i=1}^{N_{B}}\sum_{\mbox{\scriptsize$\begin{array}{c}j\in\mathbb{J}_{i}^{s}\\k\in\mathbb{K}_{i}^{s}\end{array}$}}\left\|f_{\bm{\Theta}}(\bm{x}_{i}+\bm{n}_{i}^{j})-(\bm{x}_{i}+\bm{n}_{i}^{k})\right\|_{2}^{2}\bigg ],\\
			&s.t.\quad idx(j)=idx(k),
		\end{aligned}
		\label{eq:6}
	\end{align}
\end{shrinkeq}
where $\bm{\Theta}^{s}$ is the parameters at the $s$th step, $\eta$ means the learning rate and $N_{B}$ is the batch size. $j_{l}^{s}$ and $k_{l}^{s}$ are the indexes of the input and the label randomly coupled before the $s$th step.

\begin{figure}[t]
	\centering
	\includegraphics[height=6.2cm]{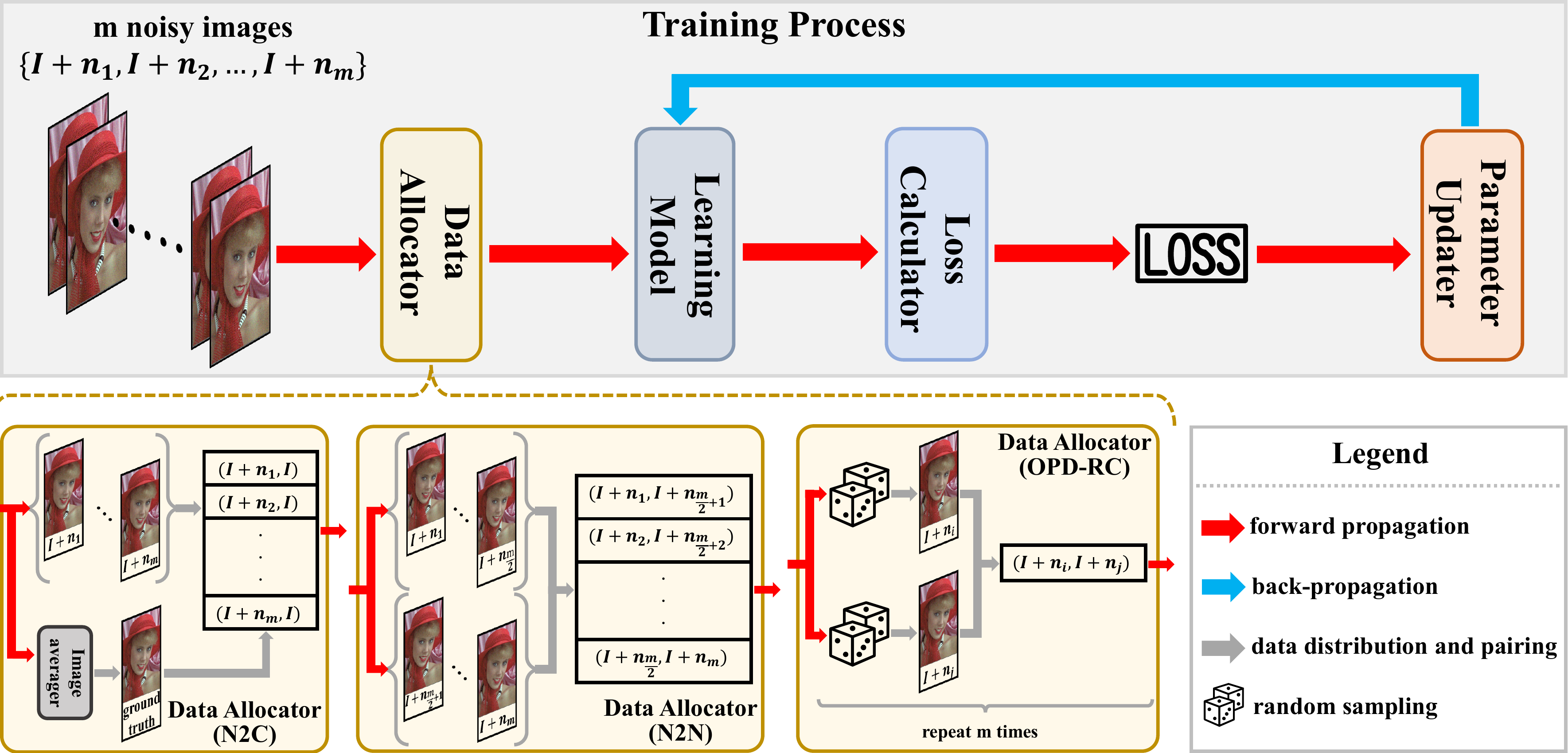}
	\vspace{-1.5ex}
	\caption{The workflows of the proposed OPD-RC and other denoising strategies including N2C and N2N. The upper part presents the common workflow to train a learning-based model. The data allocator represented by the gold box is the key difference between OPD-RC and other strategies, which are shown in detail in the three small sub-figures at the bottom.}
	\label{fig:2}
	\vspace{-1.5ex}
\end{figure}

OPD-RC makes each of the $m$ noisy frames appear in each iteration with equal probability and serve as input or label with the same chance. This operation greatly extends the diversity of data pairing and supervision without any more training time consumption. As long as the training process goes through enough iterations, it is reasonable to think that the multi-frame images are evenly used and the mutual supervision among them has been established in a practical sense. Furthermore, OPD-RC does not affect the choice and design of the network architecture and loss function. Fig. \ref{fig:2} shows the general principle of OPD-RC and its comparison with N2C and N2N. The data pairing way in the data allocator is the key difference between OPD-RC and other supervision strategies, as shown in the lower part of Fig. \ref{fig:2}.

\noindent\textbf{OPD-AL: }In order to make the $m$ noisy frames corresponding to an $\bm{x}_{i}$ play a full and balanced role during training, the second averaging operation in Eq. (\ref{eq:2}) can be moved to $f_{\bm{\Theta}}(\cdot)$:

\begin{shrinkeq}{-1.5ex}
	\begin{align}
		\mathcal{L}_{OPD}^{C}=\frac{1}{N_{B}}\sum_{i=1}^{N_{B}}\Big\|\frac{1}{m}\sum_{j=1}^{m}f_{\bm{\Theta}}(\bm{x}_{i}+\bm{n}_{i}^{j})-\widehat{\bm{x}_{i}}\Big\|_{2}^{2},
		\label{eq:7}
	\end{align}
\end{shrinkeq}
where the superscript $C$ of $\mathcal{L}_{OPD}^{C}$ indicates that this loss is still under clean supervision.

According to the polynomial theorem, reorganize $\mathcal{L}_{OPD}^{C}$:

\begin{shrinkeq}{-1.5ex}
	\begin{align}
		\mathcal{L}_{OPD}^{C}=\frac{1}{N_{B}}\sum_{i=1}^{N_{B}}\bigg[\frac{1}{m}\sum_{j=1}^{m}\left\|\bm{y}_{i}^{j}-\widehat{\bm{x}_{i}}\right\|_{2}^{2}-\frac{1}{m^{2}}\sum_{j=1}^{m-1}\sum_{k=j+1}^{m}\left\|\bm{y}_{i}^{j}-\bm{y}_{i}^{k}\right\|_{2}^{2}\bigg],
		\label{eq:8}
	\end{align}
\end{shrinkeq}
where $f_{\bm{\Theta}}(\bm{x}_{i}+\bm{n}_{i}^{j})$ and $f_{\bm{\Theta}}(\bm{x}_{i}+\bm{n}_{i}^{k})$ are replaced by $\bm{y}_{i}^{j}$ and $\bm{y}_{i}^{k}$ in writing, respectively. The specific derivation from Eq. (\ref{eq:7}) to Eq. (\ref{eq:8}) is provided in Supplementary Material (S.M).

Use mean square error (MSE) and mean square alienation (MSA) to replace the two items included in the first summation in Eq. (\ref{eq:8}):

\begin{shrinkeq}{-1.5ex}
	\begin{align}
		\mathcal{L}_{OPD}^{C}=\frac{1}{N_{B}}\sum_{i=1}^{N_{B}}\left[(\mathcal{L}_{MSE}^{C})_{i}-(\mathcal{L}_{MSA})_{i}\right]
		\label{eq:9}
	\end{align}
\end{shrinkeq}

Since Wu et al.\cite{wu2019consensus} have proved the equivalence of convergence with clean or noisy labels, replace $\widehat{\bm{x}_{i}}$ with its corresponding noisy frames, we can finally get the OPD loss:

\begin{shrinkeq}{-1.5ex}
	\begin{align}
		\mathcal{L}_{OPD}=\frac{1}{N_{B}}\sum_{i=1}^{N_{B}}\bigg[(\mathcal{L}_{MSE}^{N})_{i}-(\mathcal{L}_{MSA})_{i}\bigg],
		\label{eq:10}
	\end{align}
\end{shrinkeq}
where $(\mathcal{L}_{MSE}^{N})_{i}$ and $(\mathcal{L}_{MSA})_{i}$ are respectively formulated as:

\begin{shrinkeq}{-1.5ex}
	\begin{align}
		\begin{aligned}
			&(\mathcal{L}_{MSE}^{N})_{i}=\frac{1}{m(m-1)}\sum_{j=1}^{m}\sum_{\mbox{\scriptsize$\begin{array}{c}k=1,\\k\neq j\end{array}$}}^{m}\left\|\bm{y}_{i}^{j}-(\bm{x}_{i}+\bm{n}_{i}^{k})\right\|_{2}^{2}\\
			&(\mathcal{L}_{MSA})_{i}=\frac{1}{m^{2}}\sum_{j=1}^{m-1}\sum_{k=j+1}^{m}\left\|\bm{y}_{i}^{j}-\bm{y}_{i}^{k}\right\|_{2}^{2}
		\end{aligned}
		\label{eq:11}
	\end{align}
\end{shrinkeq}

So far, OPD loss has been successfully constructed. The key constraint on inter-frame mutual supervision is the $\mathcal{L}_{MSA}$ term, which rewards the inter-frame alienation mined by the model. When $m$ equals to $2$, Eq. (\ref{eq:11}) is reduced to the loss of N2N superimposed with mutual supervision. This shows that the proposed OPD is a generalized form of N2N.
\vspace{-1.ex}
\subsection{OPD vs. other Supervision Strategies}\label{sec:3.3}
\noindent\textbf{OPD vs. AAR: }As a learning-based strategy, OPD has no restrictions on the physical pattern of noise. However, AAR can only denoise images with zero-mean signal-independent noise. Compared with OPD, AAR is not suitable for generalization, since numerous frames under the same view are required when denoising on new data.

\noindent\textbf{OPD vs. N2C: }As an unsupervised strategy, OPD does not require any clean images as labels, which is not the case for N2C. In addition, N2C regards $m$ noisy frames corresponding to the same clean target as independent samples, whereas OPD regards them as a whole for more global consideration and more comprehensive mining.

\noindent\textbf{OPD vs. N2N: }Compared with OPD, N2N does not fully utilize multi-frame data. Pairwise matching in $m$ frames and roles for input and label are both determined arbitrarily before training. It is easy to realize that the $m$ noisy images corresponding to each $\bm{x}_{i}$ are equally valuable and should play an equal role, which is exactly what OPD achieves.
\vspace{-1.ex}
\section{Experiments}
\begin{figure}[t]
	\centering
	\includegraphics[height=6.8cm]{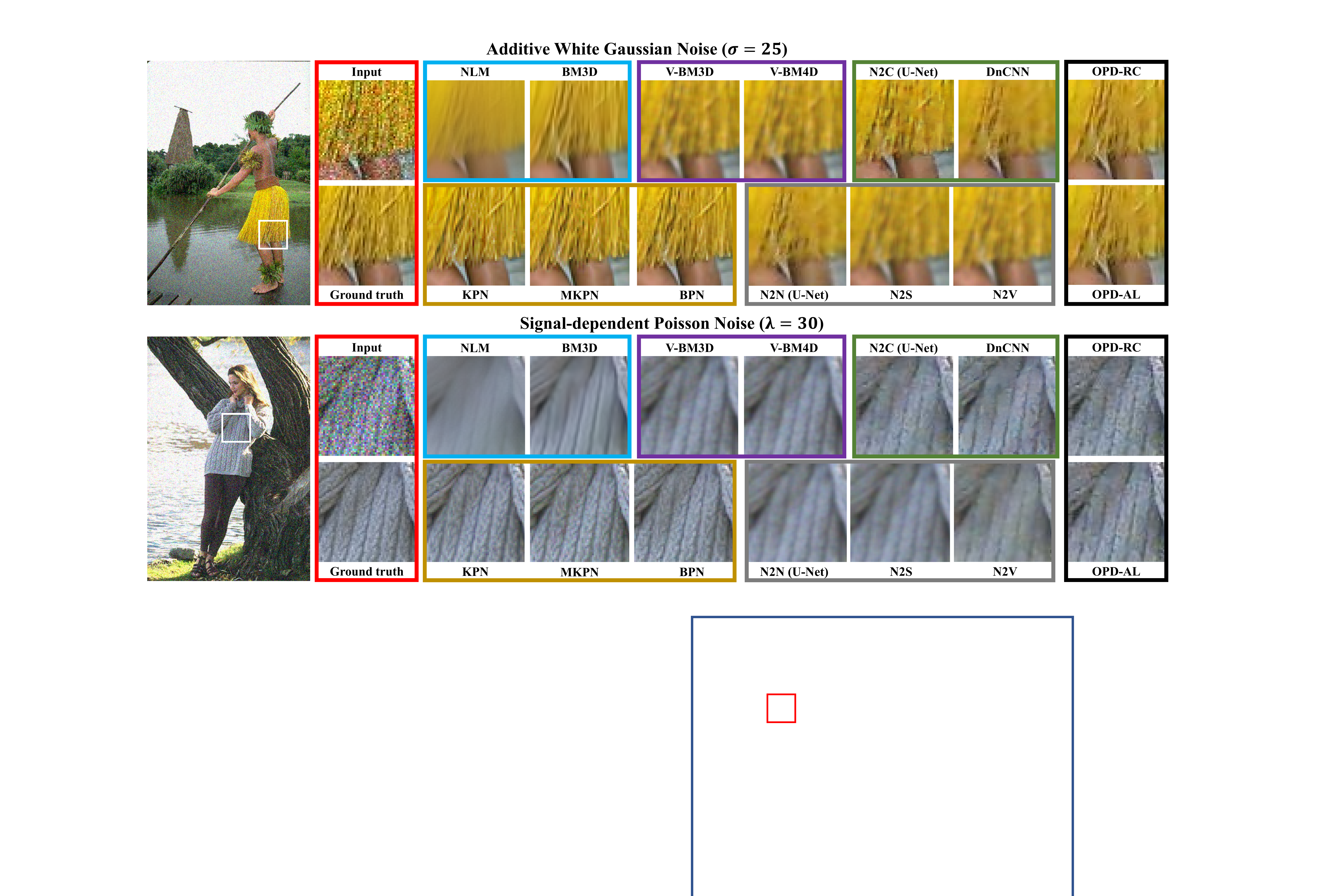}
	\caption{Example results of denoising synthetic noise. The results of different categories of methods are framed with different colored boxes. The categories are listed in Tab. \ref{tab:1}.}
	\label{fig:3}
\end{figure}

Our OPD is experimented in three typical scenarios: synthetic Gaussian and Poisson noise, and OCT speckle noise. Representatives of non-learning (e.g. NLM\cite{buades2005non}, BM3D\cite{dabov2007image}), supervised (e.g. N2C, KPN\cite{mildenhall2018burst}) and unsupervised (e.g. N2N\cite{lehtinen2018noise2noise}, N2S\cite{batson2019noise2self}) denoising methods participate in the comparison, including single-image and multi-frame algorithms. All the quantitative evaluation results in this paper are statistically significant. More results are presented in S.M, and high-resolution versions of Fig. \ref{fig:3} and Fig. \ref{fig:4} are also provided in S.M.

\vspace{-1.1ex}
\subsection{Settings}\label{sec:4.1}
\noindent\textbf{Datasets: }For synthetic noise, the clean data comes from 50,000 images in the ImageNet \cite{russakovsky2015imagenet} validation set, which are cropped into $256\times 256$. We randomly add AWGN with $\sigma =25$ or Poisson noise with $\lambda =30$ to the images and the frame number is set to 8. The same process is implemented on BSD300\cite{martin2001database}, KODAK\footnote{\url{http://r0k.us/graphics/kodak/}} and SET14\cite{zeyde2010single} to build testing sets. For OCT speckle noise reduction, the data comes from PKU37\cite{geng2022triplet}. Due to the inconsistent number of frames, we only use 16 frames per sample to align frame numbers among different samples, which means our experiments use only one-third the amount of PKU37.

\noindent\textbf{Implementation Details: }Considering the training efficiency, a modified U-Net\cite{ronneberger2015u,jiang2020comparative} was chosen to be the demonstrative model for N2C, N2N, OPD-RC and OPD-AL (See S.M for network architecture). He's method\cite{he2015delving} was used for initialization. Adam\cite{kingma2014adam} was used for parameter optimization with $L_{2}$ loss. In all experiments, one-tenth of the data was randomly split from the training set to be validation set. All our experiments were conducted based on PyTorch\cite{paszke2019pytorch}. Six NVIDIA RTX 3090 graphical cards each with 24GB memory were used. The hyperparameters for each experiment were different, which could be found in S.M. Peak signal-to-noise ratio (PSNR), structural similarity (SSIM)\cite{wang2004image}, and root-mean-square error (RMSE) were used as evaluation metrics to quantify the performance of involved methods.
\vspace{-1.ex}
\subsection{Denoising Synthetic Noise}\label{sec:4.2}
The average quantitative evaluation for the three testing sets with either Gaussian or Poisson noise are shown in Tab. \ref{tab:1}. Fig. \ref{fig:3} shows example results. Numerous representative methods are involved in the comparison. According to the supervision scheme, they are divided into non-learning, supervised and unsupervised methods. According to the image scene, they are divided into single-image methods and multi-frame methods. Next, we compare and analyze OPD and other methods of various categories.

\begin{table}[t]
	\centering
	\caption{The quantitative evaluation results of denoising synthetic noise and OCT speckle noise. For each scenario, the globally highest and second highest results are denoted as {\color{red}{\textbf{red}}} and {\color{blue}{\textbf{blue}}}, respectively. Locally for unsupervised methods, the highest and the second highest results are labeled with \textbf{\uuline{double underline}} and \textbf{\uwave{wave underline}}, respectively.}
	\label{tab:1}
	\Huge
	\resizebox{\textwidth}{!}{
		\setlength{\arrayrulewidth}{1.5pt}
		\begin{tabular}{|c|c|c|c|c|c|c|c|c|c|c|c|}
			\hline
			\multicolumn{2}{|c|}{\multirow{2}{*}{Category}} & \multirow{2}{*}{Method} & \multicolumn{3}{c|}{Gaussian} & \multicolumn{3}{c|}{Poisson} & \multicolumn{3}{c|}{OCT} \\
			\cline{4-12}
			\multicolumn{2}{|c|}{} & & PSNR & SSIM & RMSE & PSNR & SSIM & RMSE & PSNR & SSIM & RMSE \\
			\hline
			\multicolumn{2}{|c|}{Input} & & 22.72 & 0.505 & 0.074 & 21.20 & 0.469 & 0.088 & 20.35 & 0.513 & 0.096 \\
			\hline
			\multirow{4}{*}{\rotatebox{90}{{\makecell[c]{non-\\learning}}}} & \multirow{2}{*}{{\makecell[c]{single-\\image}}} & NLM\cite{buades2005non} & 24.92 & 0.670 & 0.059 & 24.96 & 0.676 & 0.058 & 26.36 & 0.600 & 0.048 \\
			\cline{3-12}
			& & BM3D\cite{dabov2007image} & 25.63 & 0.774 & 0.055 & 23.82 & 0.684 & 0.066 & 26.67 & 0.612 & 0.047 \\
			\cline{2-12}
			& \multirow{2}{*}{{\makecell[c]{multi-\\frame}}} & V-BM3D\cite{kostadin2007video} & 27.50 & 0.801 & 0.051 & 25.56 & 0.707 & 0.062 & 27.62 & 0.623 & 0.044 \\
			\cline{3-12}
			& & V-BM4D\cite{maggioni2011video} & 27.86 & 0.811 & 0.051 & 25.79 & 0.711 & 0.062 & 27.87 & 0.630 & 0.043 \\
			\hline
			\multirow{5}{*}{\rotatebox{90}{supervised}} & \multirow{2}{*}{{\makecell[c]{single-\\image}}} & N2C & 28.04 & 0.798 & 0.041 & 27.92 & 0.781 & 0.041 & 29.79 & 0.898 & 0.033 \\
			\cline{3-12}
			& & DnCNN\cite{zhang2017beyond} & 29.01 & 0.827 & 0.036 & 28.39 & 0.814 & 0.039 & 28.84 & {\color{blue}{\textbf{0.871}}} & 0.036 \\
			\cline{2-12}
			& \multirow{3}{*}{{\makecell[c]{multi-\\frame}}} & KPN\cite{mildenhall2018burst} & 32.31 & 0.917 & 0.025 & 32.28 & 0.916 & {\color{blue}{\textbf{0.025}}} & 26.68 & 0.582 & 0.047 \\
			\cline{3-12}
			& & MKPN\cite{marinvc2019multi} & {\color{blue}{\textbf{32.67}}} & {\color{blue}{\textbf{0.924}}} & {\color{blue}{\textbf{0.024}}} & {\color{blue}{\textbf{32.43}}} & {\color{blue}{\textbf{0.923}}} & {\color{blue}{\textbf{0.025}}} & 28.68 & 0.592 & 0.037 \\
			\cline{3-12}
			& & BPN\cite{xia2020basis} & {\color{red}{\textbf{33.84}}} & {\color{red}{\textbf{0.942}}} & {\color{red}{\textbf{0.021}}} & {\color{red}{\textbf{33.11}}} & {\color{red}{\textbf{0.936}}} & {\color{red}{\textbf{0.023}}} & 29.00 & 0.602 & 0.036 \\
			\hline
			\multirow{5}{*}{\rotatebox{90}{unsupervised}} & \multirow{3}{*}{{\makecell[c]{single-\\image}}} & N2N\cite{lehtinen2018noise2noise} & 27.48 & 0.787 & 0.048 & 27.28 & 0.775 & 0.044 & 28.07 & 0.817 & 0.040 \\
			\cline{3-12}
			& & N2S\cite{batson2019noise2self} & 26.88 & 0.780 & 0.049 & 27.11 & 0.760 & 0.045 & 22.23 & 0.523 & 0.089 \\
			\cline{3-12}
			& & N2V\cite{krull2019noise2void} & 26.29 & 0.772 & 0.050 & 26.95 & 0.721 & 0.046 & 21.90 & 0.518 & 0.091 \\
			\cline{2-12}
			& \multirow{2}{*}{{\makecell[c]{multi-\\frame}}} & OPD-RC & \textbf{\uwave{28.15}} & \textbf{\uwave{0.805}} & \textbf{\uwave{0.040}} & \textbf{\uuline{28.22}} & \textbf{\uwave{0.789}} & \textbf{\uuline{0.040}} & {\color{red}{\textbf{\uuline{30.69}}}} & {\color{red}{\textbf{\uuline{0.900}}}} & {\color{red}{\textbf{\uuline{0.029}}}} \\
			\cline{3-12}
			& & OPD-AL & \textbf{\uuline{28.36}} & \textbf{\uuline{0.807}} & \textbf{\uuline{0.039}} & \textbf{\uwave{28.16}} & \textbf{\uuline{0.790}} & \textbf{\uuline{0.040}} & {\color{blue}{\textbf{\uwave{30.40}}}} & {\color{blue}{\textbf{\uwave{0.871}}}} & {\color{blue}{\textbf{\uwave{0.030}}}} \\
			\hline
		\end{tabular}
	}
\end{table}

\noindent\textbf{OPD vs. other Unsupervised Methods: }
Both quantitative and qualitative results show that the proposed OPD achieves SOTA among all unsupervised methods. As can be seen from Tab. \ref{tab:1}, for Gaussian noise, compared with N2N, the PSNR of OPD-AL is improved by 0.88dB, the SSIM is improved by 0.020, and the RMSE is decreased by 0.009. Similar boosts also occur on OPD-RC and for Poisson noise. Small changes in metrics show big changes in visual perception. Fig. \ref{fig:3} shows that both OPD algorithms preserve high-frequency details better for both Gaussian and Poisson noise, such as the fringes on the hula skirt in the upper example of Fig. \ref{fig:3} and the texture of the sweater in the lower example of Fig. \ref{fig:3}.

\noindent\textbf{OPD vs. Supervised Methods: }
It is unfair to compare unsupervised OPD with supervised methods, but we still do some discussion in order to evaluate OPD more comprehensively. Tab. \ref{tab:1} shows that OPD is better than N2C but worse than all other supervised methods. However, from the visual perception in Fig. \ref{fig:3}, the denoising effect of OPD is comparable to that of N2C and DnCNN\cite{zhang2017beyond}. As an unsupervised method, this result is already satisfactory.

\noindent\textbf{OPD vs. Non-learning Methods: }
Both the quantization results given in Tab. \ref{tab:1} and the examples shown in Fig. \ref{fig:3} show that OPD exhibits unquestionable denoising advantages over non-learning methods. Looking at the example in Fig. \ref{fig:3}, it is easy to see that non-learning methods tend to suffer from oversmoothing, which is well overcome by OPD.

\noindent\textbf{OPD vs. other Multi-frame Methods: }
OPD is the first unsupervised MFD method. Of course, it is reasonable that OPD as an unsupervised method is inferior to supervised multi-frame methods. However, real-life multi-frame scenes often do not support obtaining clean labels, which is precisely the significance of our research.

\noindent\textbf{OPD vs. single-image Methods: }
Regardless of supervised or unsupervised, multi-frame methods always significantly outperform single-image methods. Specifically, OPD is better than N2N\cite{lehtinen2018noise2noise}, N2S\cite{batson2019noise2self} and N2V\cite{krull2019noise2void} on detail retention. Even more astonishing, supervised multi-frame methods such as BPN\cite{xia2020basis} are almost indistinguishable from ground truth.

\begin{figure}[t]
	\centering
	\includegraphics[height=5.1cm]{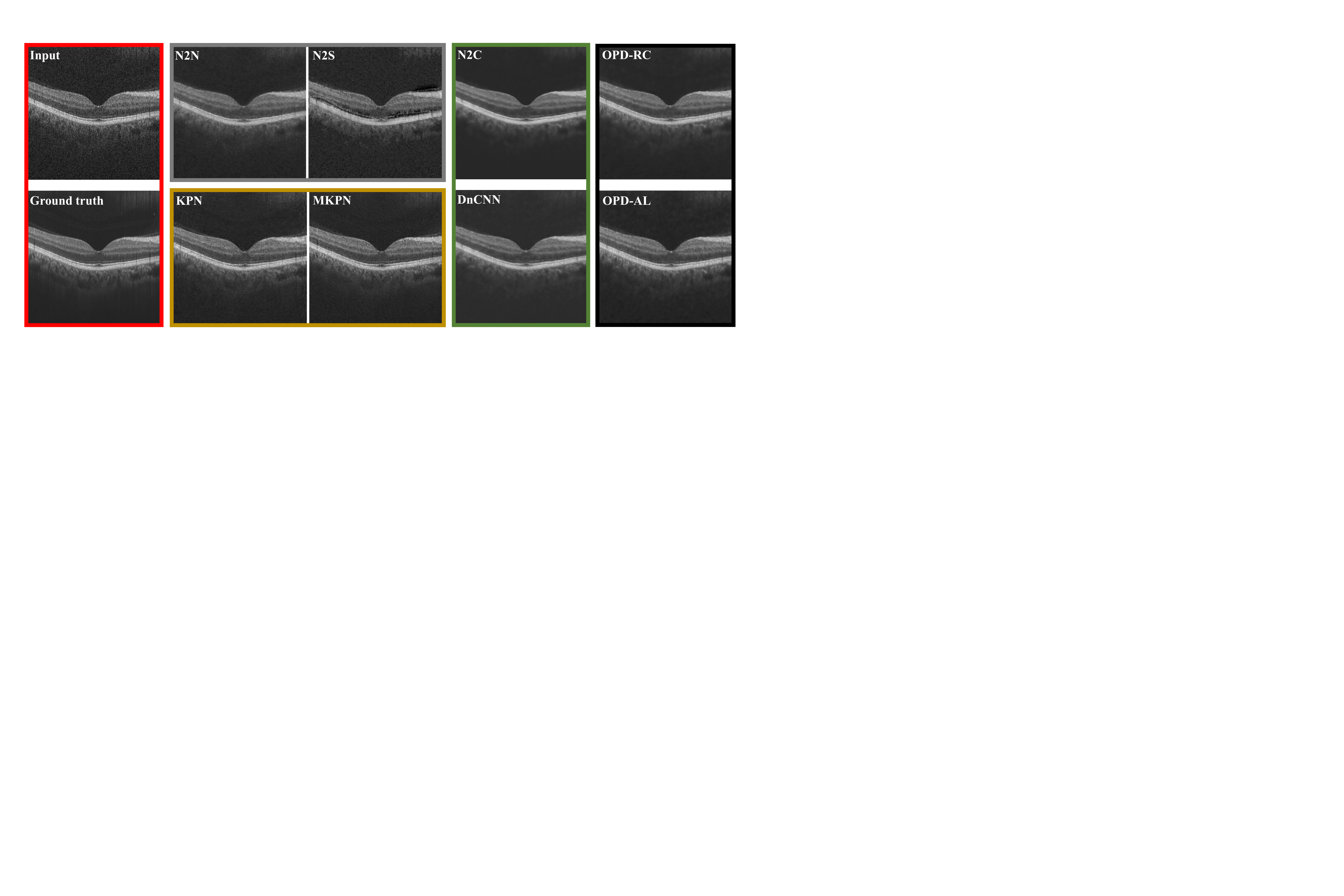}
	\caption{Example result of denoising OCT speckle noise. The results of different categories of methods are framed with different colored boxes. The categories are listed in Tab. \ref{tab:1}.}
	\label{fig:4}
	\vspace{-1.5ex}
\end{figure}
\vspace{-1.ex}
\subsection{OCT Speckle Noise Reduction}\label{sec:4.3}
The quantitative evaluation of OCT are shown in Tab. \ref{tab:1}. Fig. \ref{fig:4} shows an example result.

\noindent\textbf{OPD Wins SOTA on OCT: }Among all the methods involved in the comparison, including supervised methods, OPD-RC captures the SOTA result, and OPD-AL is only marginally behind. The two OPD methods are the only methods with PSNR exceeding 30 dB. In addition, the SSIM of OPD-RC reaches 0.9 and the RMSE reaches below 0.03. Fig. \ref{fig:4} demonstrates that the OPD-denoised image has sharper and more intact retinal layers, and the complex signal of the choroid is not over-smoothed as in other methods.

\noindent\textbf{Kernel-based Methods Fail: }Unlike what is seen on synthetic noise, supervised multi-frame methods such as KPN\cite{mildenhall2018burst} perform poorly on OCT. Destructive streaks appear in the images corresponding to KPN\cite{mildenhall2018burst}, MKPN\cite{marinvc2019multi} and BPN\cite{xia2020basis} in Fig. \ref{fig:4}. This is because the laser coherent noise contained in OCT is long-range\cite{schmitt1999speckle}, and methods based on local kernel prediction cannot mine such global noise well, but OPD based on mutual supervision can.

\noindent\textbf{Self-supervised Methods Fail: }N2S\cite{batson2019noise2self} and N2V\cite{krull2019noise2void} hardly converge on the OCT denoising task because their premise for local mask estimation is that the noise contained in the image is signal-independent, which is the opposite of the case of OCT.

\noindent\textbf{OPD-RC vs. OPD-AL: }Quantitatively, OPD-AL outperforms OPD-RC on synthetic noise, but vice versa on OCT. Qualitatively, OPD-AL-processed images seem to be sharper than those OPD-RC-processed, both on synthetic noise and OCT. This reflects that the alienation loss may be slightly better than simple randomization. In addition, the results also illustrate the importance of visual perception beyond quantitative evaluation.
\vspace{-1.ex}
\section{Conclusion}
\vspace{-1.ex}
For the first time, our work defines the concept of mutual supervision and proposes an unsupervised strategy named OPD for MFD. Unlike pairwise supervision in traditional learning strategies, OPD uniformly establishes supervision relationships among multiple images participating in learning. We propose two specific algorithms, OPD-RC and OPD-AL, respectively from the perspectives of data allocation and alienation loss design. The experiments show the effectiveness of our OPD strategy on several MFD tasks including denoising Gaussian and Poisson noise and OCT speckle noise reduction.
\vspace{-1.ex}
\section*{Acknowledgments}
\vspace{-1.ex}
This work was supported in part by the Beijing Natural Science Foundation (Z210008) and in part by the Shenzhen Science and Technology Program (KQTD20180412181221912, JCYJ20200109140603831).

\bibliography{egbib}



\section*{Supplementary material}

Supplementary Material for
# One-Pot Multi-Frame Denoising

Lujia Jin[123]
jinlujia@pku.edu.cn
Shi Zhao[4]
magishe@pku.edu.cn
Lei Zhu[123]
zhulei@stu.pku.edu.cn
Qian Chen[123]
chen_qian@stu.pku.edu.cn
Yanye Lu[✉235]
yanye.lu@pku.edu.cn

[1] Department of Biomedical Engineering,
 Peking University, Beijing, China
[2] Institute of Medical Technology,
 Peking University, Beijing, China
[3] Institute of Biomedical Engineering,
 Peking University Shenzhen Graduate
 School, Shenzhen, China
[4] School of Physics,
 Peking University, Beijing, China
[5] National Biomedical Imaging Center,
 Peking University, Beijing, China

**Abstract**

This document is a supplement to the BMVC 2022 submission # 0061: "One-Pot Multi-Frame Denoising". The supplementary material covers OPD-related theoretical derivation, algorithm implementation, and experimental results. Sec. 1 provides a rigorous derivation of the alienation loss in the OPD-AL method. Sec. 2 gives additional details about the experimental implementation, including the network architecture used for demonstration (2.1), hyperparameter settings for each experiment (2.2) and the explanation and selection of evaluation metrics (2.3). In Sec. 3, we conduct a series of ablation studies to explore the relationship between the proposed OPD strategy and some factors including mutuality of supervision (3.1), image multiplicity (3.2), noise level (3.3), and training efficiency (3.4). Sec. 4 provides additional experimental results containing high-resolution version of the example results in the main paper (4.1), respective results on BSD300, KODAK and SET14 (4.2), and other example results for denoising AWGN and Poisson noise (4.3).

## 1 Rigorous Derivation of Alienation Loss

We explain the derivation of $\mathcal{L}_{OPD}$, which refines the process from the Eq. (7) to the Eq. (8) in the main paper.

Let us first review the polynomial theorem.

**Theorem 1** *Any nonnegative power of a polynomial summation can be expanded into a sum of the form:*

$$(a_1 + a_2 + ... + a_p)^q = \sum_{l_1+l_2+...+l_p=q} \frac{q!}{l_1! l_2! ... l_p!} a_1^{l_1} a_2^{l_2} ... a_p^{l_p} \qquad (1)$$

*where $n, m \in \mathbb{N}$, $n, m \geq 2$ and $k_i \in [0, n]$, $i \in [1, m]$.*

The Eq. (7) in the main paper is formulated as:

$$\mathcal{L}_{OPD}^{C} = \frac{1}{N_B} \sum_{i=1}^{N_B} \mathcal{L}_i = \frac{1}{N_B} \sum_{i=1}^{N_B} \left\| \frac{1}{m} \sum_{j=1}^{m} f_{\Theta}(\boldsymbol{x}_i + \boldsymbol{n}_i^j) - \widehat{\boldsymbol{x}}_i \right\|_2^2, \qquad (2)$$

where the superscript $C$ of $\mathcal{L}_{OPD}^{C}$ indicates that this loss is still under clean supervision. Replace the $f_{\Theta}(\boldsymbol{x}_i + \boldsymbol{n}_i^j)$ by $\boldsymbol{y}_i^j$ and use the way described in Theorem 1 to disassemble the summation:

$$\begin{aligned}
\mathcal{L}_i &= \frac{1}{m^2} \left\| \sum_{j=1}^{m} (\boldsymbol{y}_i^j - \widehat{\boldsymbol{x}}_i) \right\|_2^2 \\
&= \frac{1}{m^2} \sum_{j=1}^{m} \left\| \boldsymbol{y}_i^j - \widehat{\boldsymbol{x}}_i \right\|_2^2 + \frac{1}{m^2} \sum_{\substack{j,k=1,\\k \neq j}}^{m} (\boldsymbol{y}_i^j - \widehat{\boldsymbol{x}}_i) \cdot (\boldsymbol{y}_i^k - \widehat{\boldsymbol{x}}_i) \\
&= \mathcal{L}_{i1} + \mathcal{L}_{i2},
\end{aligned} \qquad (3)$$

where $\mathcal{L}_{i1}$ and $\mathcal{L}_{i2}$ respectively refer to the two terms in the summation. Further decompose $\mathcal{L}_{i2}$:

$$\begin{aligned}
\mathcal{L}_{i2} &= \frac{1}{m^2} \sum_{\substack{j,k=1,\\k \neq j}}^{m} (\boldsymbol{y}_i^j - \widehat{\boldsymbol{x}}_i) \cdot (\boldsymbol{y}_i^k - \widehat{\boldsymbol{x}}_i) \\
&= \frac{1}{m^2} \sum_{\substack{j,k=1,\\k \neq j}}^{m} \boldsymbol{y}_i^j \cdot \boldsymbol{y}_i^k - \frac{2(m-1)}{m^2} \sum_{j=1}^{m} \widehat{\boldsymbol{x}}_i \cdot \boldsymbol{y}_i^j + \frac{m-1}{m} \left\| \widehat{\boldsymbol{x}}_i \right\|_2^2 \\
&= \frac{m-1}{m^2} \left[ \sum_{j=1}^{m} \left\| \boldsymbol{y}_i^j \right\|_2^2 - 2 \sum_{j=1}^{m} \widehat{\boldsymbol{x}}_i \cdot \boldsymbol{y}_i^j + m \left\| \widehat{\boldsymbol{x}}_i \right\|^2 \right] - \frac{m-1}{m^2} \sum_{j=1}^{m} \left\| \boldsymbol{y}_i^j \right\|_2^2 \\
&\quad + \frac{1}{m^2} \sum_{\substack{j,k=1,\\k \neq j}}^{m} \boldsymbol{y}_i^j \cdot \boldsymbol{y}_i^k \\
&= \frac{m-1}{m^2} \sum_{j=1}^{m} \left\| \boldsymbol{y}_i^j - \widehat{\boldsymbol{x}}_i \right\|_2^2 - \frac{1}{m^2} \left[ (m-1) \sum_{j=1}^{m} \left\| \boldsymbol{y}_i^j \right\|_2^2 - 2 \sum_{j=1}^{m-1} \sum_{k=j+1}^{m} \boldsymbol{y}_i^j \cdot \boldsymbol{y}_i^k \right] \\
&= \frac{m-1}{m^2} \sum_{j=1}^{m} \left\| \boldsymbol{y}_i^j - \widehat{\boldsymbol{x}}_i \right\|_2^2 - \frac{1}{m^2} \sum_{j=1}^{m-1} \sum_{k=j+1}^{m} \left\| \boldsymbol{y}_i^j - \boldsymbol{y}_i^k \right\|_2^2
\end{aligned} \qquad (4)$$

Put the result of Eq. (4) into Eq. (3):

$$\begin{aligned}
\mathcal{L}_i &= \frac{1}{m^2} \sum_{j=1}^{m} \left\| \boldsymbol{y}_i^j - \widehat{\boldsymbol{x}}_i \right\|_2^2 + \frac{m-1}{m^2} \sum_{j=1}^{m} \left\| \boldsymbol{y}_i^j - \widehat{\boldsymbol{x}}_i \right\|_2^2 - \frac{1}{m^2} \sum_{j=1}^{m-1} \sum_{k=j+1}^{m} \left\| \boldsymbol{y}_i^j - \boldsymbol{y}_i^k \right\|_2^2 \\
&= \frac{1}{m} \sum_{j=1}^{m} \left\| \boldsymbol{y}_i^j - \widehat{\boldsymbol{x}}_i \right\|_2^2 - \frac{1}{m^2} \sum_{j=1}^{m-1} \sum_{k=j+1}^{m} \left\| \boldsymbol{y}_i^j - \boldsymbol{y}_i^k \right\|_2^2
\end{aligned} \qquad (5)$$

Put the result of Eq. (5) into Eq. (2), then we can get $\mathcal{L}_{OPD}^{C}$ shown by the Eq. (8) in the main paper:

$$\mathcal{L}_{OPD}^{C} = \frac{1}{N_B} \sum_{i=1}^{N_B} \left[ \frac{1}{m} \sum_{j=1}^{m} \left\| \boldsymbol{y}_i^j - \widehat{\boldsymbol{x}}_i \right\|_2^2 - \frac{1}{m^2} \sum_{j=1}^{m-1} \sum_{k=j+1}^{m} \left\| \boldsymbol{y}_i^j - \boldsymbol{y}_i^k \right\|_2^2 \right] \qquad (6)$$

## 2 Supplementary Details of the Experiments

Due to space limitations, the main paper only provides necessary instructions for the experimental implementation. This section gives more supplementary details including network architecture (2.1), hyperparameter settings (2.2) and evaluation metrics (2.3).

### 2.1 Network Architecture

Considering training as efficient as possible without impairing model performance too much, a modified U-Net[12, 22] is selected as the demonstrative model. U-Net is first proposed by Ronneberger et al.[22] and achieves promising performance in image segmentation tasks. Since then, many variants of U-Net have been designed and perform prominently in various tasks[4, 6, 7, 9, 20, 24]. Due to the high efficiency of training and hierarchical feature concatenation, U-Net and its variants are widely used in tasks related to image restoration and reconstruction, especially in medical image scenes, such as computed tomography (CT)[8], optical coherence tomography (OCT)[21], and positron emission tomography (PET)[23].

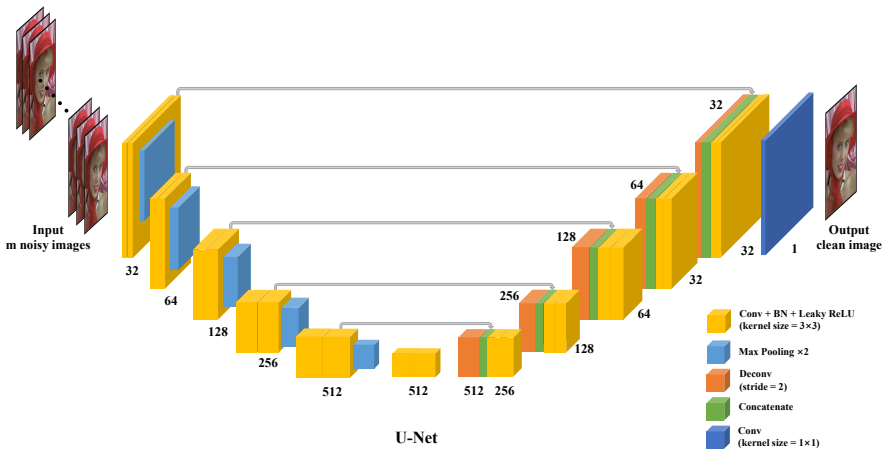

Figure 1: Schematic diagram of the structure of the U-Net we used. (Conv: convolutional layer; BN: batch normalization layer; Deconv: deconvolution layer.)

Fig. 1 demonstrates the structure of the U-Net we used, including contracting encoder and expanding decoder. The encoder reduces the size of the feature maps hierarchically while increasing the number of channels. Conversely, the decoder hierarchically enlarges the size of the feature maps while reducing the number of channels. Both the encoder and decoder contain five stages. Each stage of the encoder performs three steps programmatically, which are two $3 \times 3$ convolutions followed by batch normalization (BN)[10] and Leaky ReLU[16], and a $2 \times 2$ max pooling with a stride of 2. Each stage of the decoder performs four steps programmatically, which are a $2 \times 2$ deconvolution with a stride of 2, a concatenation with the symmetrical feature map in the encoder, and two consecutive convolutions followed by BN and Leaky ReLU. The number of parameters of our model is about 15 million.

### 2.2 Hyperparameter Settings

When training the denoising models separately for three different noises, the hyperparameter settings are obviously different. The batch size (BS) and the initial learning rate (iLR) set-

tings for the four strategies of N2C, N2N[15], OPD-RC and OPD-AL on the three scenarios are shown in Table 1. For the four strategies mentioned above, during training, the learning rate is reduced by half every 10 epochs and the number of epochs is set to 300. For other learning-based methods participating in the comparison, such as KPN[19], model training is performed according to the settings provided by the corresponding papers. Hyperparameters used in experiments corresponding to the OPD-RC results presented in the main paper are not specifically optimized but use the same hyperparameters as the experiments under N2C and N2N, which allows us to better see that, other things being equal, the proposed simple RC operation magically give the model a performance boost. We also specially optimized the hyperparameters for OPD-RC, which are given in Tab. 2. Tab. 2 shows that the performance of OPD can be further slightly improved with hyperparamenters optimization.

Table 1: The hyperparameter settings for the experiments on the synthetic noise (Gaussian & Poisson), OCT and OCTA datasets under N2C, N2N, OPL-RC and OPL-LD strategies. (BS: batch size; iLR: initial learning rate.)

|  | OPD-RC / N2C / N2N | | OPD-AL | |
|---|---|---|---|---|
|  | BS | iLR ($\times 10^{-3}$) | BS | iLR ($\times 10^{-3}$) |
| AWGN | 64 | 7.2 | 4 | 1.8 |
| Poisson noise | 64 | 4.8 | 4 | 1.2 |
| OCT | 4 | 1.8 | 1 | 0.9 |

Table 2: Hyperparameters optimization for OPD-RC.

|  | hyperparameters | | | | denoising results | | | | | |
|---|---|---|---|---|---|---|---|---|---|---|
|  | BS | | iLR ($\times 10^{-3}$) | | PSNR | | SSIM | | RMSE | |
| optimization | w/ | w/o | w/ | w/o | w/ | w/o | w/ | w/o | w/ | w/o |
| AWGN | 64 | 64 | 5.6 | 7.2 | 28.28 | 28.15 | 0.807 | 0.805 | 0.040 | 0.040 |
| Poisson | 64 | 64 | 3.2 | 4.8 | 28.30 | 28.22 | 0.790 | 0.789 | 0.039 | 0.040 |
| OCT | 4 | 4 | 1.0 | 1.8 | 30.71 | 30.69 | 0.902 | 0.900 | 0.028 | 0.029 |

## 2.3 Evaluation Metrics

Peak signal-to-noise ratio (PSNR), structural similarity (SSIM)[25], and root-mean-square error (RMSE) are used as evaluation metrics to quantify the performance of the involved methods. The essence of these evaluations is quantitative comparison between the predicted denoised image $\widehat{x}$ and the clean reference image $x$.

PSNR measures the global distortion and noise level between two images by calculating error at the pixel level. The larger the PSNR, the better the quality of the predicted image. The calculation formula of PSNR is:

$$PSNR(x,\widehat{x}) = 10\log \frac{MAX_x^2}{MSE(x,\widehat{x})}, \quad (7)$$

where $MAX_x$ represents the maximum pixel value in the ground truth image. $MSE(x,\widehat{x})$ refers to the mean square error between $\widehat{x}$ and $x$, which is calculated as follows:

$$MSE(x,\widehat{x}) = \frac{\sum_{i=1}^{H}\sum_{j=1}^{W}(\widehat{x}_{i,j} - x_{i,j})^2}{H \times W}, \quad (8)$$

where $H$ and $W$ are the height and the width of the image respectively.

SSIM measures the structural deviation between two images by comprehensively considering the information of brightness, contrast, and structure. Therefore, SSIM is sensitive to differences in local structure and the contrast of images, which is similar to human visual perception. The calculation formula of SSIM is:

$$SSIM(\bm{x},\widehat{\bm{x}}) = \frac{(2\mu_{\bm{x}}\mu_{\widehat{\bm{x}}} + C_1)(2\sigma_{\bm{x},\widehat{\bm{x}}} + C_2)}{(\mu_{\bm{x}}^2 + \mu_{\widehat{\bm{x}}}^2 + C_1)(\sigma_{\bm{x}}^2 + \sigma_{\widehat{\bm{x}}}^2 + C_2)}, \qquad (9)$$

where $\mu_{\bm{x}}$ and $\mu_{\widehat{\bm{x}}}$ represent the mean value of $\bm{x}$ and $\widehat{\bm{x}}$ respectively, and $\sigma_{\bm{x}}$ and $\sigma_{\widehat{\bm{x}}}$ represent the standard deviation of $\bm{x}$ and $\widehat{\bm{x}}$ respectively. $\sigma_{\bm{x},\widehat{\bm{x}}}$ refers to the covariance between $\bm{x}$ and $\widehat{\bm{x}}$. $C_1$ and $C_2$ are constant terms used to maintain numerical stability. In our experiments, $C_1$ and $C_2$ are respectively set to $1 \times 10^{-4}$ and $9 \times 10^{-4}$, which are their most frequently used empirical values.

RMSE represents the square root of the second moment of the difference between the pixel values of the predicted image and the reference image. The calculation formula of RMSE is:

$$RMSE(\bm{x},\widehat{\bm{x}}) = \sqrt{\frac{\sum_{i=1}^{H}\sum_{j=1}^{W}(\widehat{\bm{x}}_{i,j} - \bm{x}_{i,j})^2}{H \times W}}, \qquad (10)$$

where $H$ and $W$ are the height and the width of the image respectively.

RMSE is always non-negative, and in general, a low RMSE is better than a high RMSE. Unlike PSNR and SSIM, RMSE is extremely sensitive to outliers, which enables the failure of local image denoising to be more accurately quantified.

## 3 Ablation Study

The main paper has demonstrated the strong performance of the OPD strategy on multi-frame denoising. However, the exploration of the many possible underlying factors behind this superior performance of OPD and the robustness of OPD in different situations is still lacking. This section provides additional discussions on these issues, including the importance of the mutuality of supervision (3.1) and image multiplicity (3.2) for OPD, the robustness of OPD to various noise levels (3.3), and the impact of OPD on training efficiency (3.4).

### 3.1 Does the Mutuality of Supervision really Help?

We hold the view that the outstanding performance of the proposed OPD stems from mutual supervision. In order to verify this point of view more practically, we reduce the number of frames to 2. At this time, mutual supervision means the random exchange of the sample and the label within data pairs during training, which is called N2N with a shuffle probability of 0.5 ($N2N_{S0.5}$). From the quantitative results given in Tab. 3, it can be seen that $N2N_{S0.5}$ outperforms N2N in all aspects. Additionally, we simply learn data pairs in two supervision directions one after another, which is named bidirectional learning (BDL). This twice-augmented dataset performs almost as well as $N2N_{S0.5}$. However, the rough augmentation of the dataset consumes almost twice of training time that under $N2N_{S0.5}$. Furthermore, we conduct gradient experiments on the OCT dataset with probabilities ranging from 0 to 1. The results (Fig. 2) also show that the mutuality really boosts the performance of OPD on multi-frame denoising.

### 3.2 Does a Higher Image Multiplicity really Help?

The greater the number of frames, the greater the number of mutual-supervision combinations. Gradient experiments were carried out on OCT and the results are shown in Fig. 3.

Table 3: Evaluation of the effectiveness and high training efficiency of mutual supervision. $N2N_{S0.5}$ means N2N with a shuffle probability of 0.5 and BDL means bidirectional learning. The best results are marked in **red**.

|  | AWGN | | | | Poisson noise | | | | OCT | | | |
|---|---|---|---|---|---|---|---|---|---|---|---|---|
|  | PSNR | SSIM | RMSE | T(s)/Iter. | PSNR | SSIM | RMSE | T(s)/Iter. | PSNR | SSIM | RMSE | T(s)/Iter. |
| N2N | 27.48 | 0.787 | 0.048 | **419.7** | 27.28 | 0.775 | 0.044 | **420.1** | 28.07 | 0.817 | 0.040 | **25.9** |
| $N2N_{S0.5}$ | **27.58** | **0.801** | **0.043** | 422.1 | 27.79 | **0.779** | **0.042** | 421.6 | **28.63** | **0.864** | **0.037** | 26.3 |
| BDL | 27.56 | 0.798 | 0.044 | 780.7 | **27.80** | **0.779** | **0.042** | 779.2 | 28.56 | 0.863 | **0.037** | 49.6 |

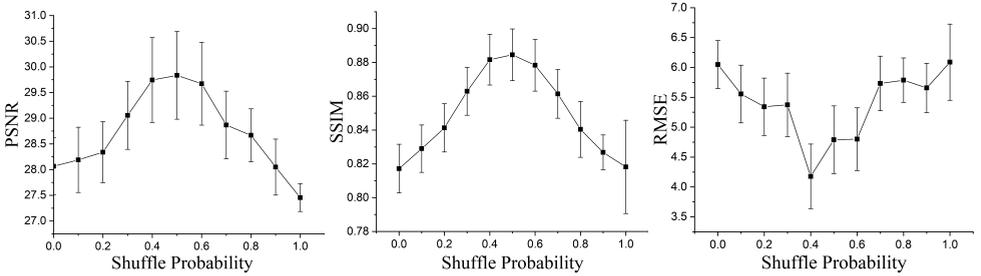

Figure 2: OCT speckle noise reduction using $N2N_S$ with different shuffle probabilities.

To enhance comparability and facilitate visualization, only the four curves corresponding to N2C, N2N, OPD-RC and OPD-AL are drawn in Fig. 3 and other methods are not considered in this comparison. Additionally, for the numerical discrimination of the RMSE for elegant visualization, the RMSE is calculated using the image with pixel values in the range [0, 255] in Fig. 3 so that its values lie between 0 and 10. It can be seen that, with the increase of image multiplicity, the superiority of OPD in comparison with N2N and N2C gradually reduces. On the one hand, the skyrocketing amount of data might gradually bridge the difference between methods. On the other hand, the theoretically best denoising performance on multi-frame denoising is proportional to the square root of the image multiplicity $m$[3], which makes it less and less profitable as the number of frames growing.

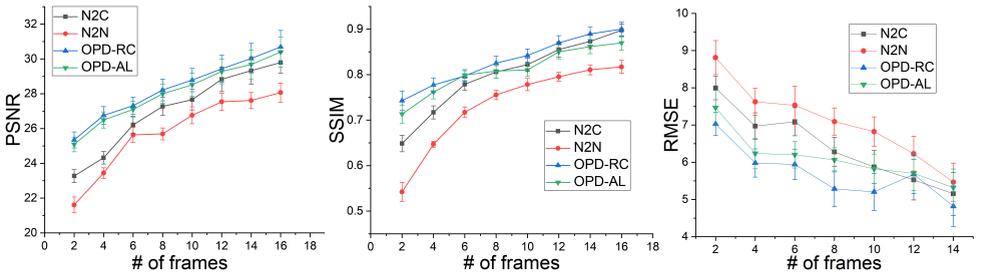

Figure 3: OCT speckle noise reduction using OPD and other supervision strategies with gradient image multiplicity.

## 3.3  Is OPD Robust to Various Noise Levels?

In the denoising experiments for synthetic noise in the main paper, $\sigma$ for Gaussian noise is set to 25 and $\lambda$ for Poisson noise is set to 30. This setting corresponds to the most common levels of both types of noise in the real world[11]. However, the robustness of the proposed OPD strategy to more faint or intense levels of noise deserves further study. Therefore, we

performed gradient experiments. The σ of the Gaussian noise is separately set to 5 gradients from 5 to 45, with a difference of 10 between each two adjacent gradients. The λ of Poisson noise is separately set to 5 gradients from 10 to 50, and the difference between every two adjacent gradients is also 10. The results of the denoising experiments are shown in Fig. 4. In order to facilitate comparison and display, only five curves corresponding to Input, N2C, N2N, OPD-RC and OPD-AL are given in Fig. 4, and other methods are not involved in the comparison. Additionally, for the numerical discrimination of the RMSE for elegant visualization, the RMSE is calculated using the image with pixel values in the range [0, 255] in Fig. 3 so that its values lie between 0 and 10. It can be seen from Fig. 4 that both OPD-RC and OPD-AL perform stronger than N2N and comparable to N2C for any level of noise, no matter for Gaussian or Poisson noise and no matter for which of the three metrics. These results fully reflects the robustness of OPD to different levels of noise contained in images.

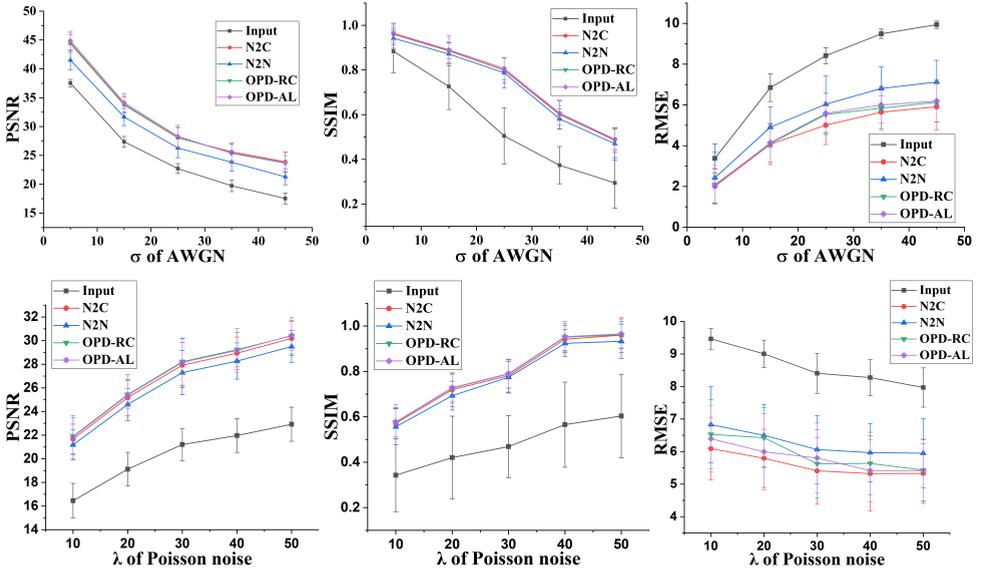

Figure 4: Evaluation of denoising results for different levels of Gaussian and Poisson noise using OPD and other supervised strategies.

## 3.4 Can OPD Improve the Efficiency of Training?

In principle, the OPD strategy only improves the diversity of supervision by increasing the potential space of paired data, and does not actually increase the amount of training data, so there is no theoretical obstacle to the convergence of model training. To verify this point, we recorded and plotted in Fig. 5 the loss curves for different strategies during training of the model for denoising AWGN. Because OPD-AL uses its own loss function, which makes the model iteration process different from other strategies without strict variable constraints, OPD-AL is not included in the comparison here. For the clarity and beauty of the visualization, only N2C and N2N are compared with OPD-RC in Fig. 5, and the loss curves on the training set and validation set are displayed separately in two subplots. It is evident from Fig. 5 that, instead of slowing down the model convergence, OPD-RC significantly accelerates the convergence process compared to N2N and N2C. This result is surprising but reasonable, because OPD improves the diversity of supervision and brings additional space for optimiza-

tion paths for the model to search and optimize in the parameter space. From this point of view, the proposed OPD is not only effective but also efficient.

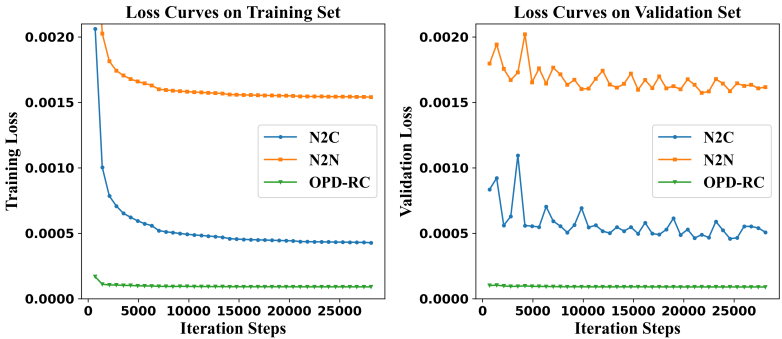

Figure 5: Loss curves during training for denoising AWGN.

## 4 Additional Results

Due to space limitations, only a few representative experimental results are given in the main paper, which will be supplemented in this section. Sec. 4.1 gives high-resolution versions of the example results corresponding to Fig. 3 and 4 in the main paper. The quantitative evaluation results of synthetic noise in the main paper are averaged on the three test sets of BSD300, KODAK and SET14, and Sec. 4.2 gives the respective results on them. In addition, Sec. 4.2 also provides two additional example results for BSD300, KODAK and SET14 each, including one for AWGN and the other for Poisson noise. Sec. 4.3 provides additional example results of denoising AWGN and Poisson noise.

### 4.1 High-resolution Version of Example Results in the Paper

Fig. 3 and 4 in the main paper give representative examples of experimental results for AWGN, Poisson noise and OCT speckle noise. In order to better show the details, Fig 6, 7 and 8 respectively give their corresponding high-resolution versions for comparing and viewing. For better presentation, the results for AWGN and Poisson noise are shown separately in Fig. 6 and 7.

### 4.2 Respective Results on BSD300, KODAK and SET14

The main paper only provides average quantitative evaluation results on three test sets of BSD300, KODAK and SET14 for synthetic AWGN and Poisson noise. The respective results on the three datasets are given in Tab. 4.

### 4.3 Additional Example Results for AWGN and Poisson Noise

Qualitative results for denoising AWGN and Poisson noise are supplemented in this section. Fig. 9, 10 and 11 give example results for BSD300, KODAK and SET14 for AWGN, respectively. Fig. 12, 13 and 14 give example results for BSD300, KODAK and SET14 for Poisson noise, respectively.

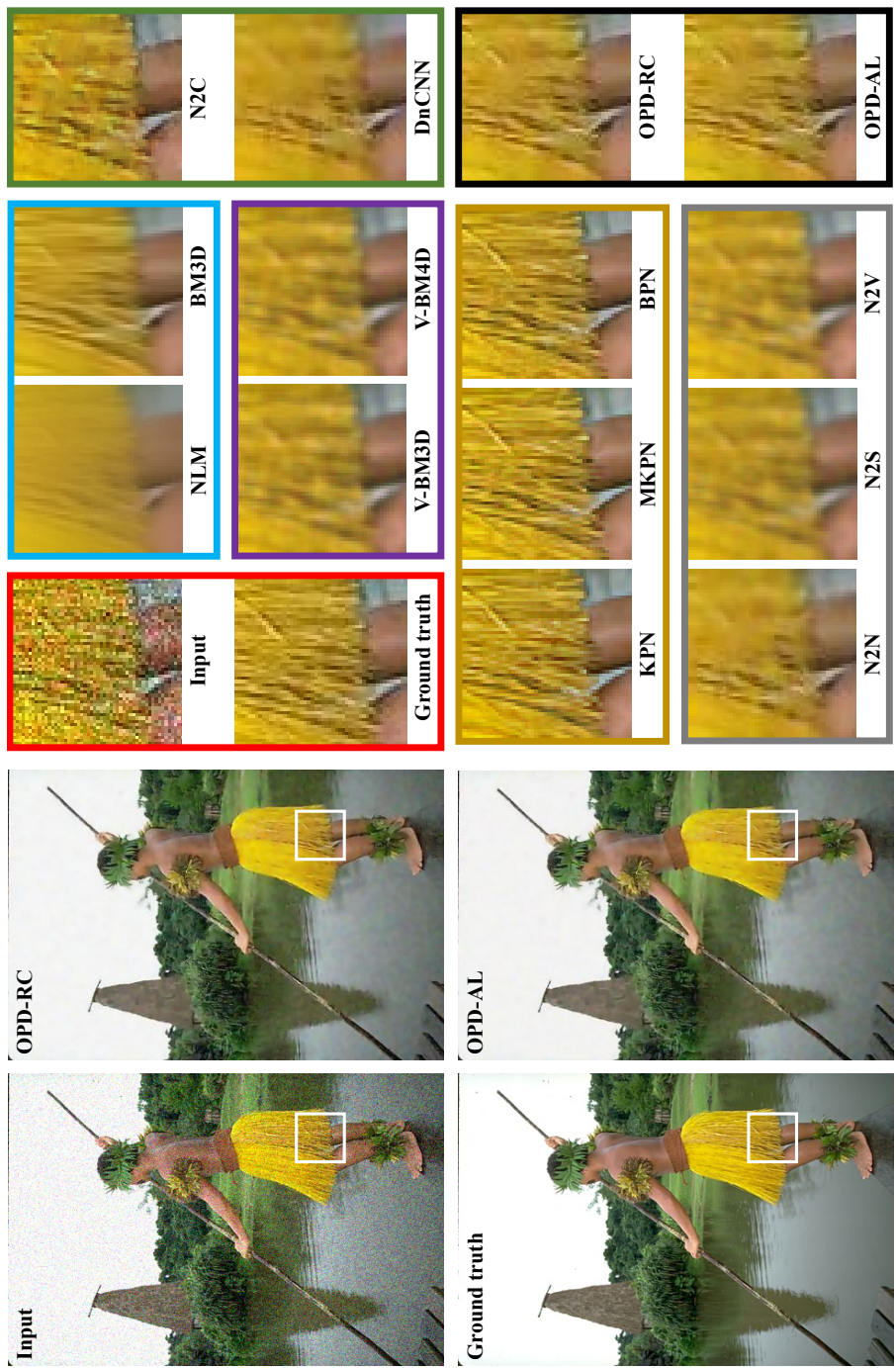

Figure 6: The high-resolution version of the upper part of the Fig. 3 in the main paper (an example result for denoising AWGN).

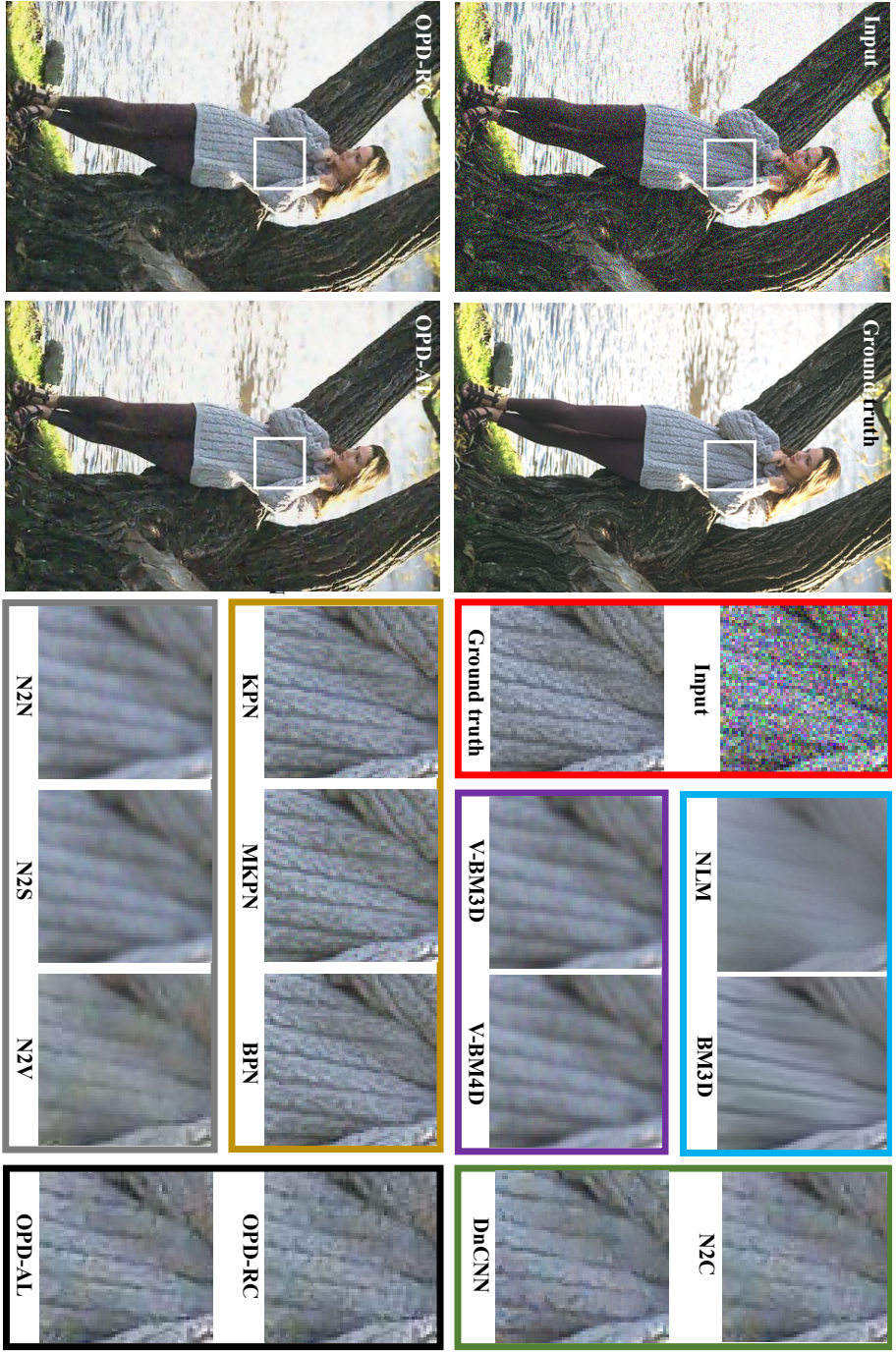

Figure 7: The high-resolution version of the lower part of the Fig. 3 in the main paper (an example result for denoising Poisson noise).

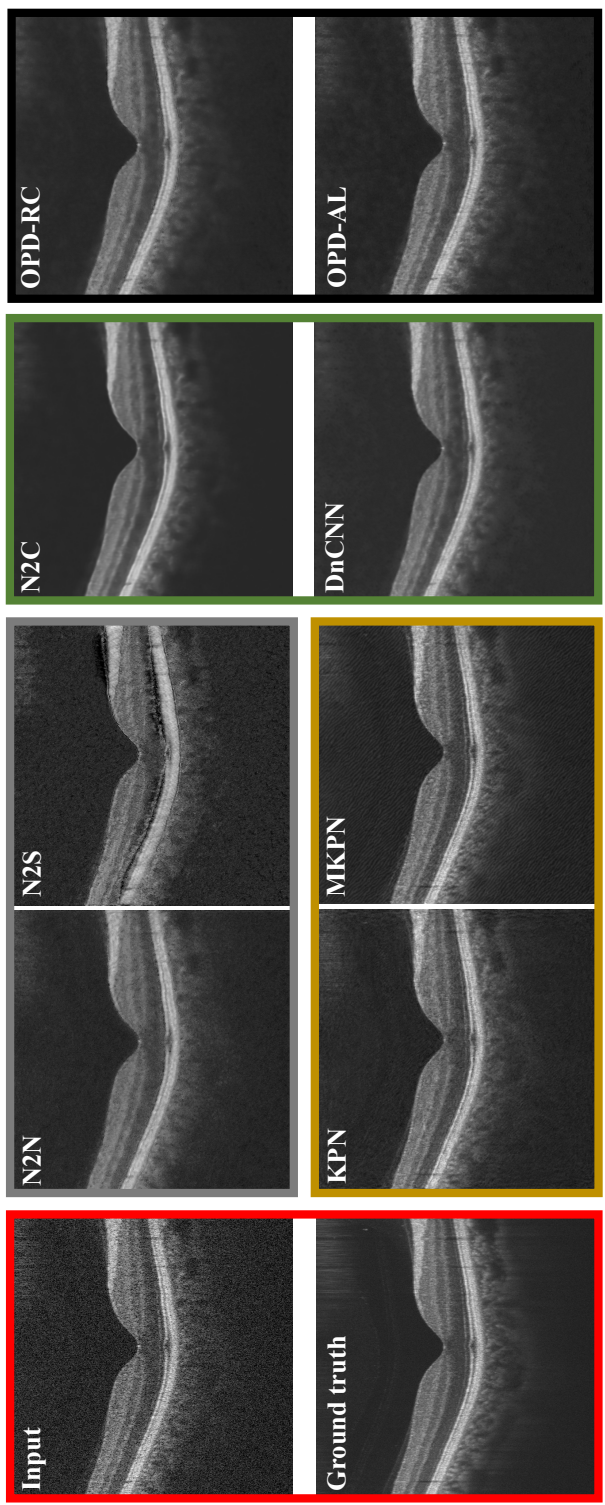

Figure 8: The high-resolution version of the Fig. 4 in the main paper (an example result for OCT denoising).

Table 4: The quantitative evaluation results of denoising AWGN on BSD300, KODAK and SET14. For each dataset, the globally highest and second highest results are denoted as **red** and **blue**, respectively. Locally for unsupervised methods, the highest and the second highest results are labeled with **double underline** and ~~wave underline~~.

| Category | | Input | Method | BSD300 | | | KODAK | | | SET14 | | | average | | |
|---|---|---|---|---|---|---|---|---|---|---|---|---|---|---|---|
| | | | | PSNR | SSIM | RMSE | PSNR | SSIM | RMSE | PSNR | SSIM | RMSE | PSNR | SSIM | RMSE |
| non-learning | single-image | | NLM[2] | 23.00 | 0.492 | 0.071 | 20.44 | 0.611 | 0.095 | 20.60 | 0.595 | 0.093 | 22.72 | 0.505 | 0.074 |
| | | | BM3D[5] | 24.73 | 0.667 | 0.060 | 26.75 | 0.697 | 0.047 | 25.74 | 0.688 | 0.053 | 24.92 | 0.670 | 0.059 |
| | multi-frame | | V-BM3D[13] | 25.64 | 0.773 | 0.055 | 26.08 | 0.785 | 0.051 | 24.71 | 0.779 | 0.062 | 25.63 | 0.774 | 0.055 |
| | | | V-BM4D[17] | 27.45 | 0.793 | 0.051 | 28.35 | 0.853 | 0.049 | 27.37 | 0.840 | 0.052 | 27.50 | 0.801 | 0.051 |
| supervised | single-image | | N2C | 27.82 | 0.799 | 0.051 | 28.53 | 0.869 | 0.048 | 27.62 | 0.832 | 0.051 | 27.86 | 0.811 | 0.051 |
| | | | DnCNN[27] | 28.00 | 0.791 | 0.041 | 28.79 | 0.867 | 0.037 | 27.72 | 0.845 | 0.042 | 28.04 | 0.798 | 0.041 |
| | multi-frame | | KPN[19] | 28.64 | 0.828 | 0.037 | 30.08 | 0.822 | 0.032 | 28.47 | 0.814 | 0.039 | 29.01 | 0.827 | 0.036 |
| | | | MKPN[18] | 32.21 | 0.918 | 0.025 | 33.83 | 0.916 | 0.021 | 31.91 | 0.898 | 0.027 | 32.31 | 0.917 | 0.025 |
| | | | BPN[26] | **32.68** | **0.923** | **0.024** | **33.83** | **0.924** | **0.021** | **32.38** | **0.915** | **0.025** | **32.67** | **0.924** | **0.024** |
| unsupervised | single-image | | **33.76** | **0.943** | **0.021** | **35.07** | **0.941** | **0.018** | **33.21** | **0.924** | **0.023** | **33.84** | **0.942** | **0.021** |
| | | | N2N[15] | 27.41 | 0.779 | 0.048 | 28.58 | 0.864 | 0.042 | 26.99 | 0.836 | 0.049 | 27.48 | 0.787 | 0.048 |
| | | | N2S[1] | 26.86 | 0.771 | 0.050 | 28.06 | 0.862 | 0.043 | 26.43 | 0.830 | 0.051 | 26.88 | 0.780 | 0.049 |
| | | | N2V[14] | 26.22 | 0.763 | 0.050 | 27.51 | 0.860 | 0.043 | 25.89 | 0.821 | 0.052 | 26.29 | 0.772 | 0.050 |
| | multi-frame | | OPD-RC | **28.07** | **0.797** | **0.040** | **29.25** | **0.882** | **0.035** | **27.92** | **0.850** | **0.041** | **28.15** | **0.805** | **0.040** |
| | | | OPD-AL | <u>28.28</u> | <u>0.800</u> | <u>0.040</u> | <u>29.42</u> | <u>0.875</u> | <u>0.034</u> | <u>28.10</u> | <u>0.852</u> | <u>0.040</u> | <u>28.36</u> | <u>0.807</u> | <u>0.039</u> |

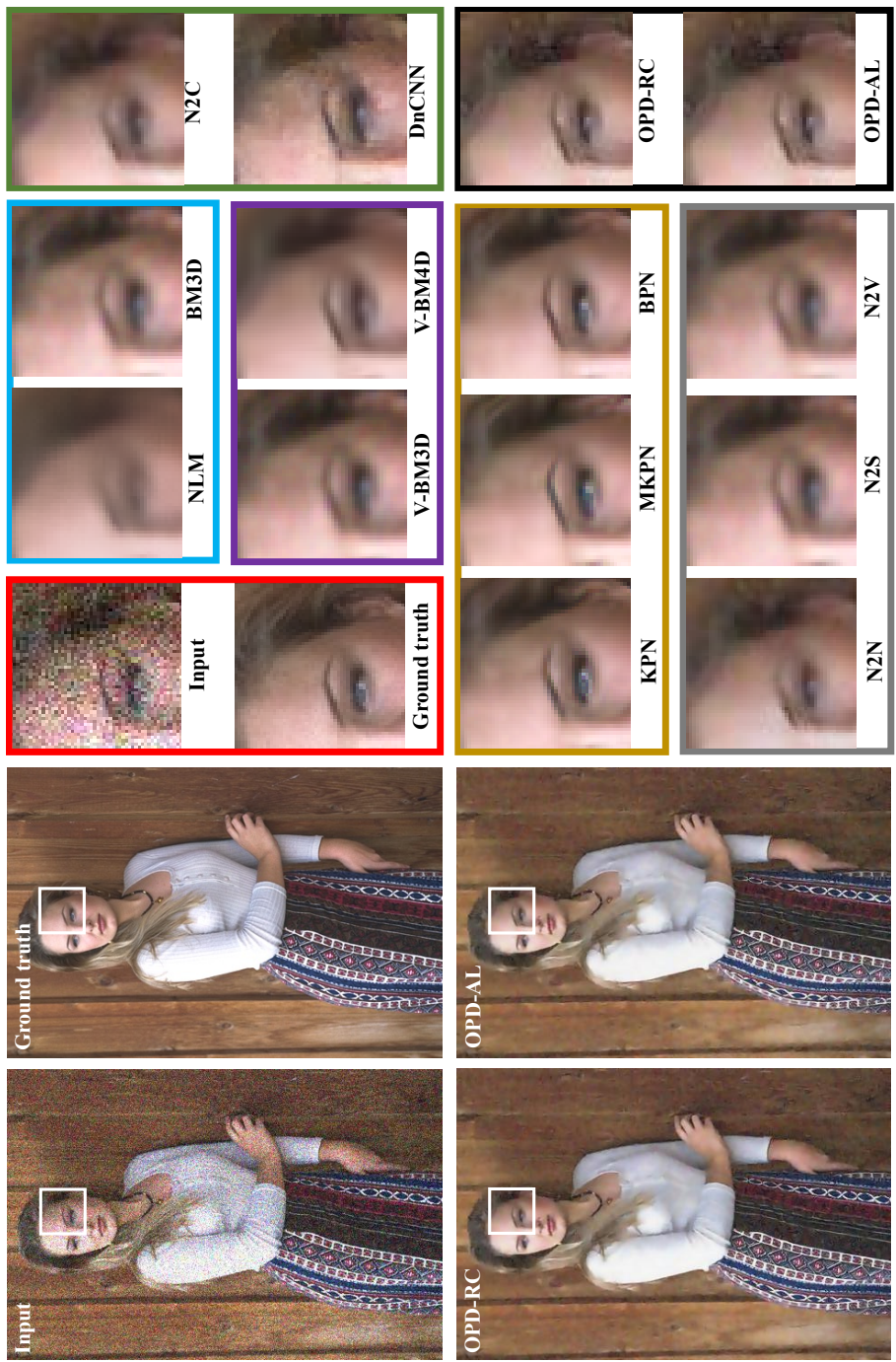

Figure 9: An example result from BSD300 for denoising AWGN.

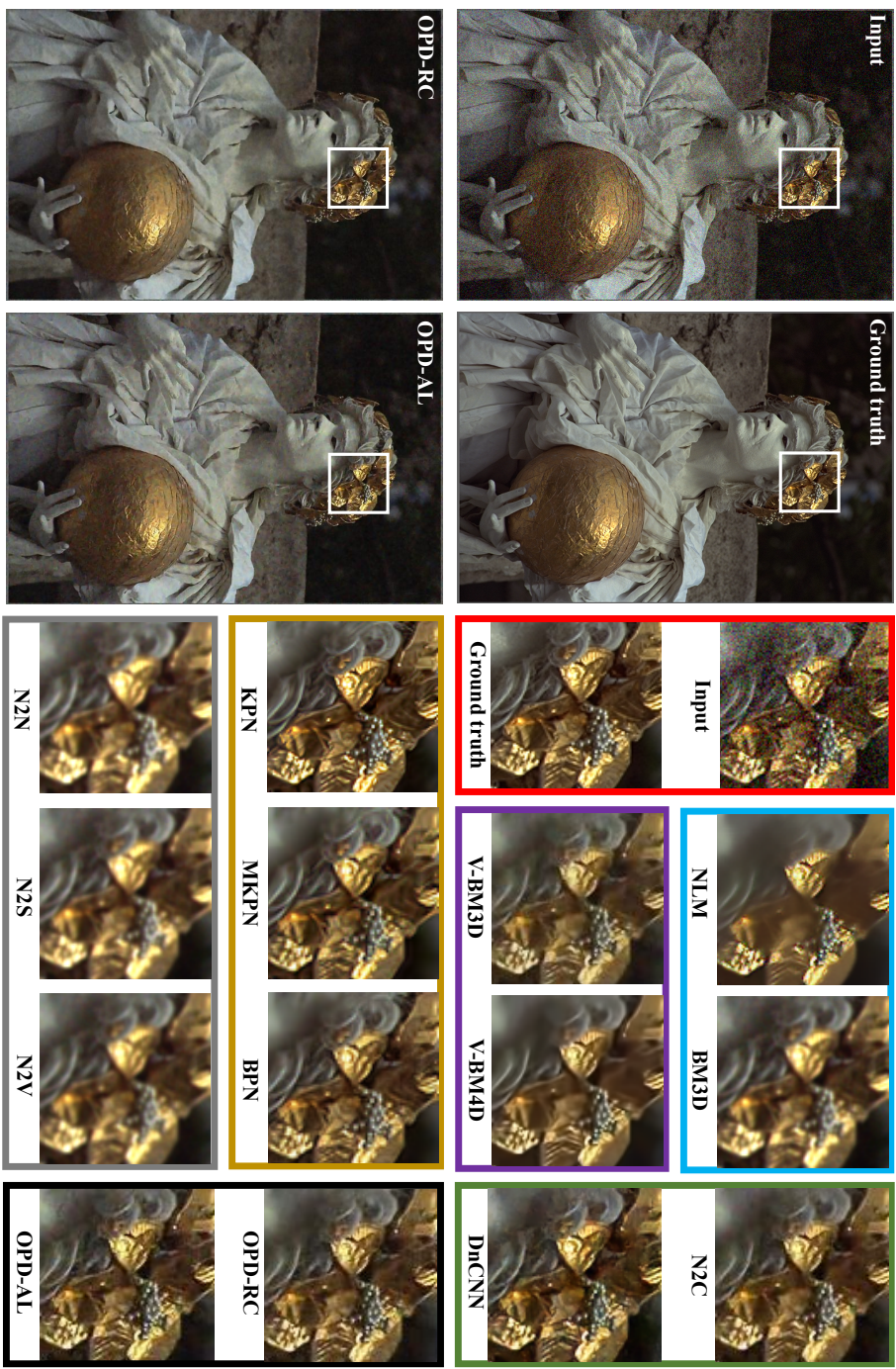

Figure 10: An example result from KODAK for denoising AWGN.

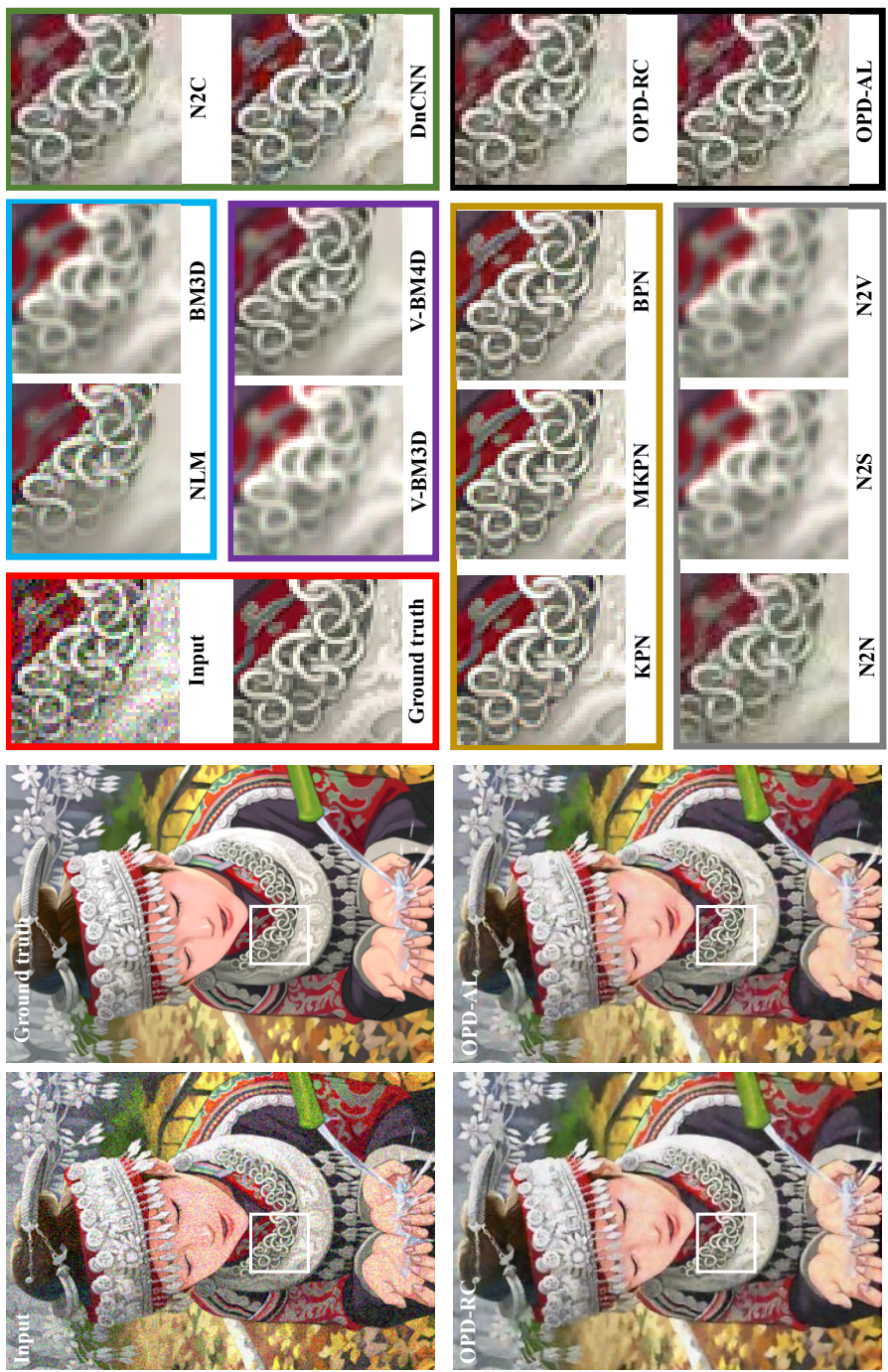

Figure 11: An example result from SET14 for denoising AWGN.

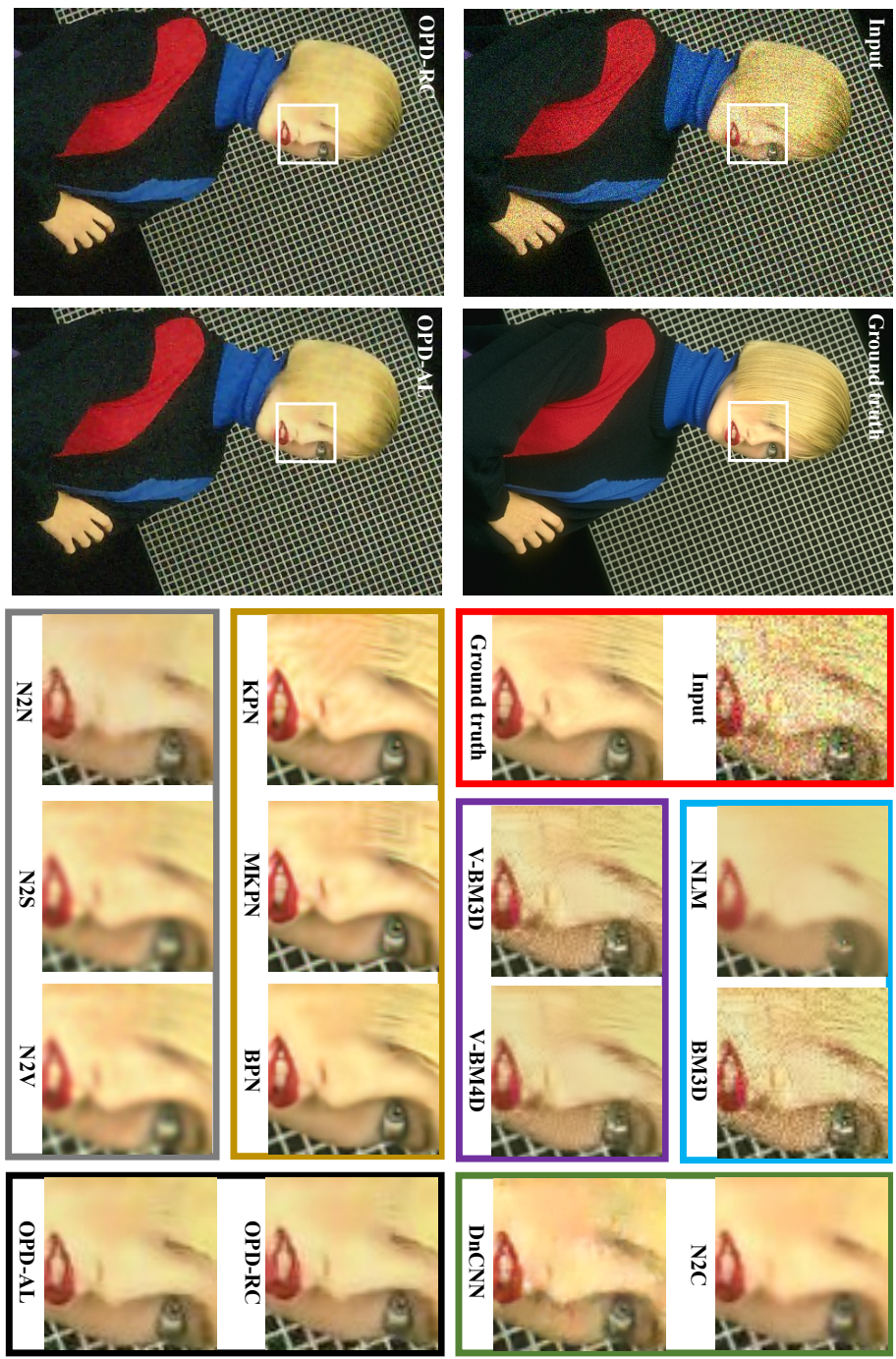

Figure 12: An example result from BSD300 for denoising Poisson noise.

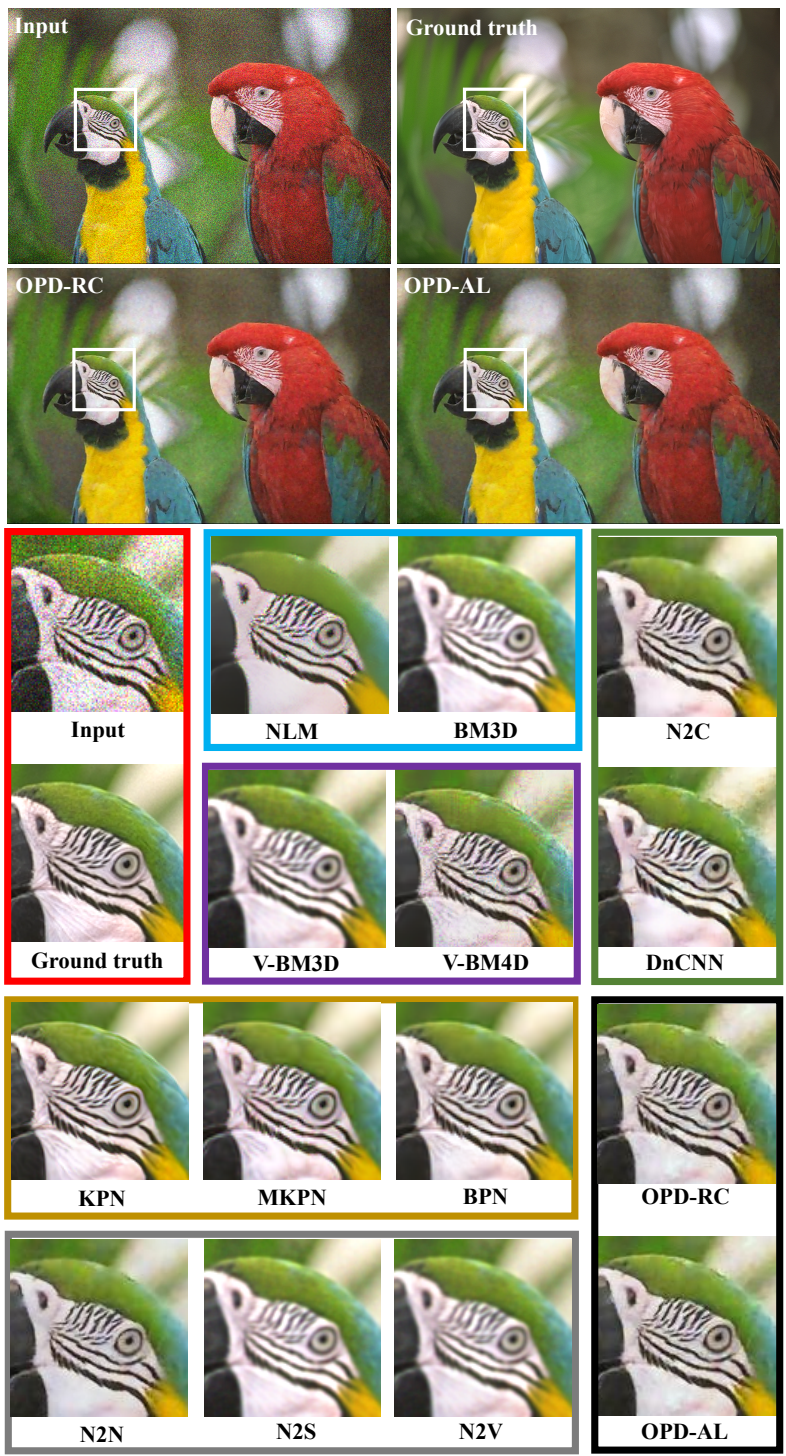

Figure 13: An example result from KODAK for denoising Poisson noise.

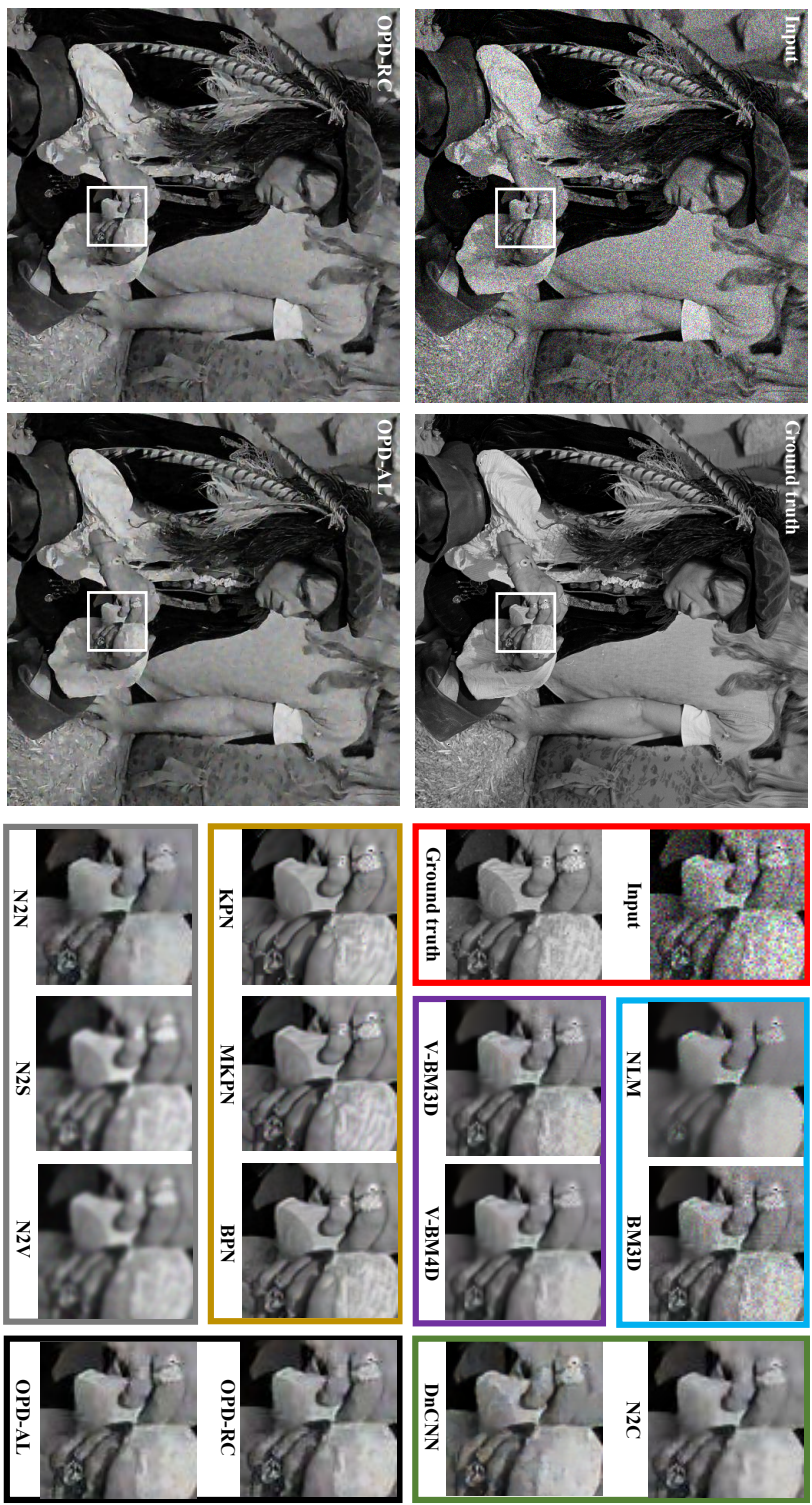

Figure 14: An example result from SET14 for denoising Poisson noise.



\end{document}